\documentclass[10pt,twocolumn,letterpaper]{article}

\usepackage[pagenumbers]{cvpr} % To force page numbers, e.g. for an arXiv version

\usepackage{graphicx}
\usepackage{amsmath}
\usepackage{amssymb}
\usepackage{booktabs}
\usepackage{placeins}
\usepackage{multirow}
\usepackage{enumitem}
\usepackage{wrapfig}
\usepackage{makecell}
\usepackage{tablefootnote}

\usepackage[accsupp]{axessibility}

\usepackage{xcolor}         % colors

\usepackage{amsthm}
\theoremstyle{definition}
\newtheorem{defn}{Definition}
\DeclareMathOperator{\argmin}{argmin} % no space, limits on side in displays

\usepackage[pagebackref,breaklinks,colorlinks]{hyperref}

\usepackage[capitalize]{cleveref}
\crefname{section}{Sec.}{Secs.}
\Crefname{section}{Section}{Sections}
\Crefname{table}{Table}{Tables}
\crefname{table}{Tab.}{Tabs.}

\def\cvprPaperID{6896} % *** Enter the CVPR Paper ID here
\def\confName{CVPR}
\def\confYear{2023}

\begin{document}

\title{The Best Defense is a Good Offense: Adversarial Augmentation against Adversarial Attacks}

\author{Iuri Frosio, NVIDIA\\ {\tt\small ifrosio@nvidia.com}
\and
Jan Kautz, NVIDIA\\ {\tt\small jkautz@nvidia.com}
}
\maketitle

%%%%%%%%% ABSTRACT

%%%%%%%%% ABSTRACT
\begin{abstract}
  Many defenses against adversarial attacks (\eg robust classifiers, randomization, or image purification) use countermeasures put to work only after the attack has been crafted.
  We adopt a different perspective to introduce $A^5$ (Adversarial Augmentation Against Adversarial Attacks), a novel framework including the first certified preemptive defense against adversarial attacks.
  The main idea is to craft a defensive perturbation to guarantee that any attack (up to a given magnitude) towards the input in hand will fail. 
  To this aim, we leverage existing automatic perturbation analysis tools for neural networks.
  We study the conditions to apply $A^5$ effectively, analyze the importance of the robustness of the to-be-defended classifier, and inspect the appearance of the robustified images.
  We show effective on-the-fly defensive augmentation with a robustifier network that ignores the ground truth label, and demonstrate the benefits of  robustifier and classifier co-training.
  In our tests, $A^5$ consistently beats state of the art certified defenses on MNIST, CIFAR10, FashionMNIST and Tinyimagenet.
  We also show how to apply $A^5$ to create certifiably robust physical objects.
  Our code at  \url{https://github.com/NVlabs/A5} allows experimenting on a wide range of scenarios beyond the man-in-the-middle attack tested here, including the case of physical attacks.
\end{abstract}

%%%%%%%%% BODY TEXT

\section{Introduction}
\label{sec:introduction}

\begin{figure}[b!]
\centering
\vspace{-0.3cm}
\begin{tabular}{cc}
\includegraphics[width=0.14\textwidth,trim=2.6cm 0.4cm 1.1cm 1.5cm, clip]{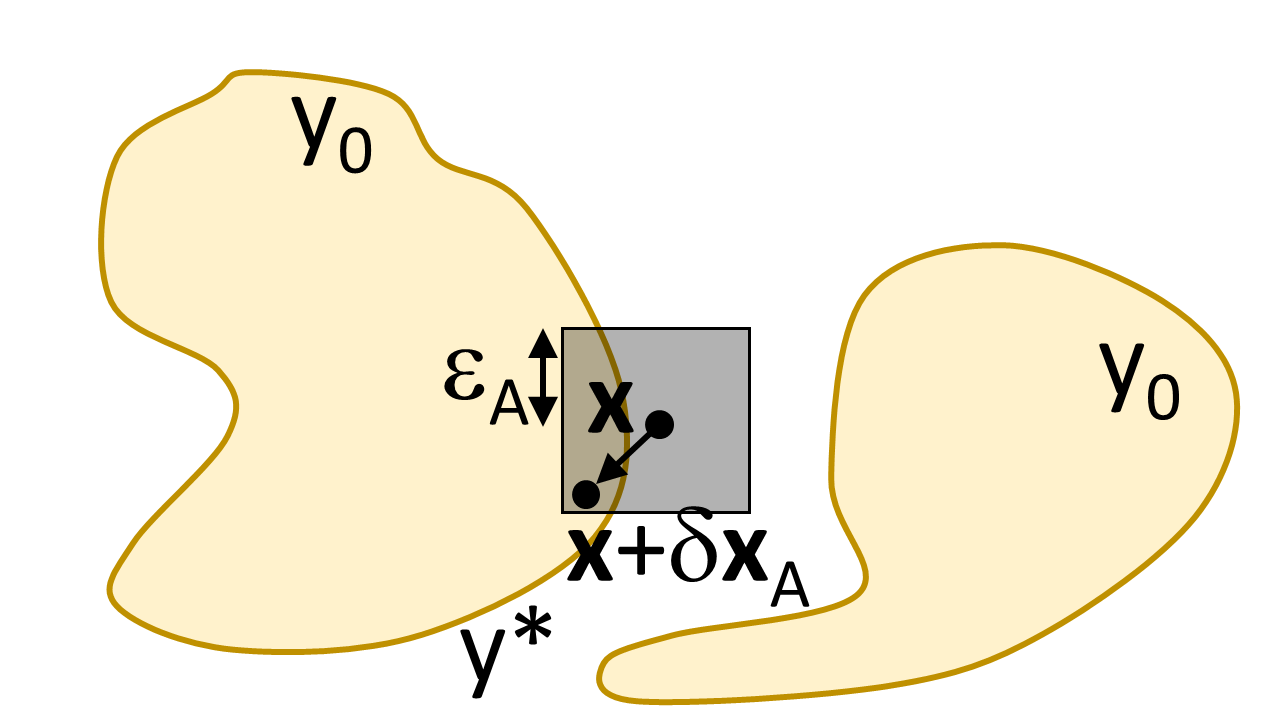} &
\includegraphics[width=0.14\textwidth,trim=2.6cm 0.4cm 1.1cm 1.5cm, clip]{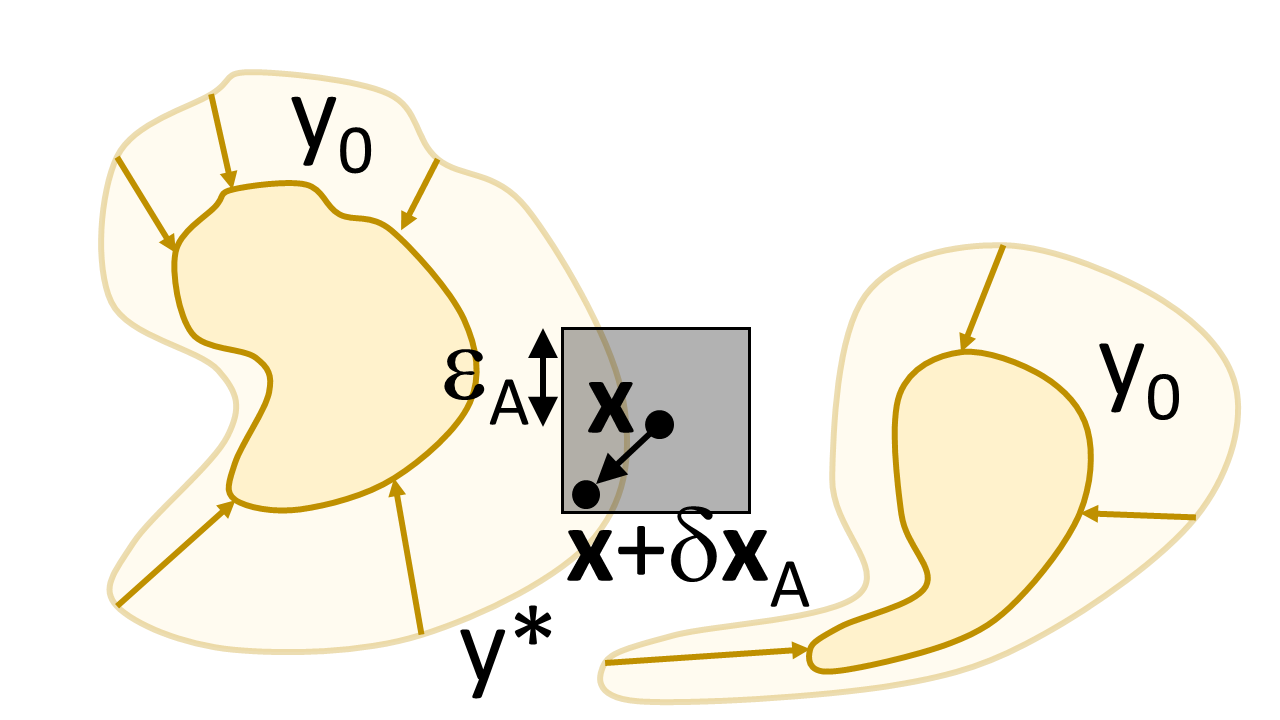} \\
(a) certified bounds & (b) adversarial training \vspace{0.1cm} \\
%\hline
\includegraphics[width=0.14\textwidth,trim=2.6cm 0.4cm 1.1cm 1.2cm, clip]{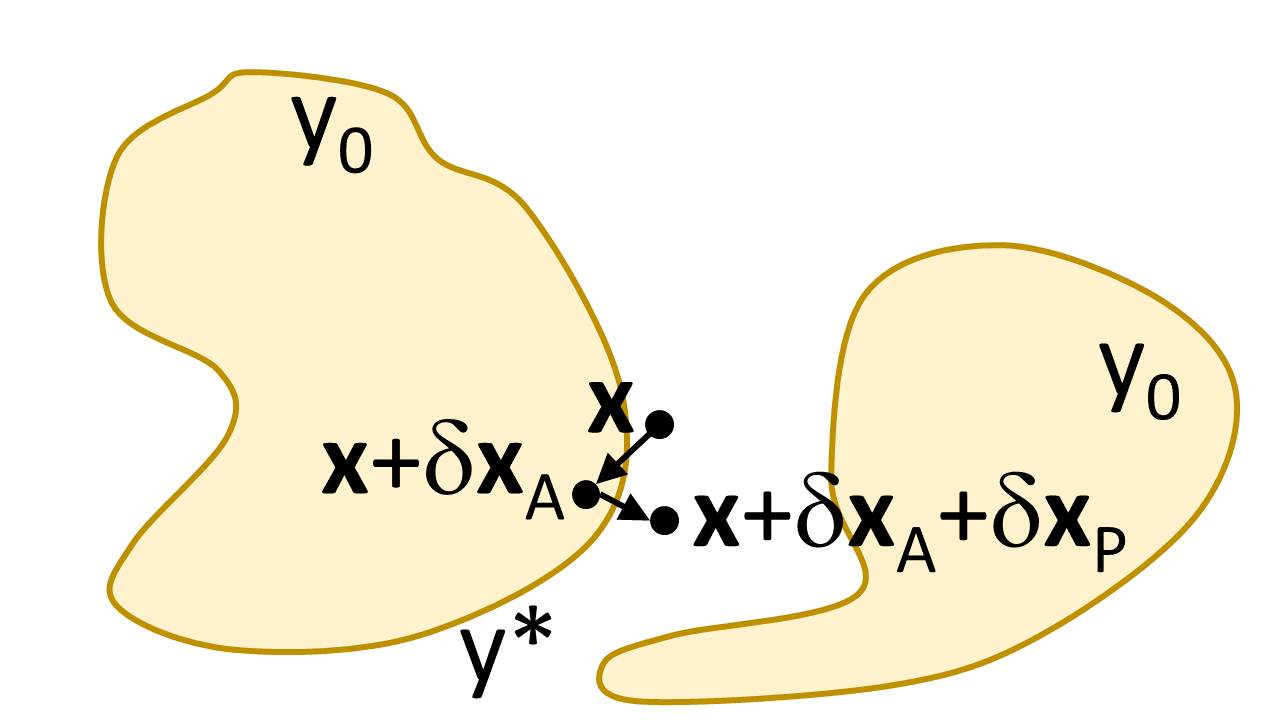} & 
\includegraphics[width=0.14\textwidth,trim=2.6cm 0.4cm 1.1cm 1.2cm, clip]{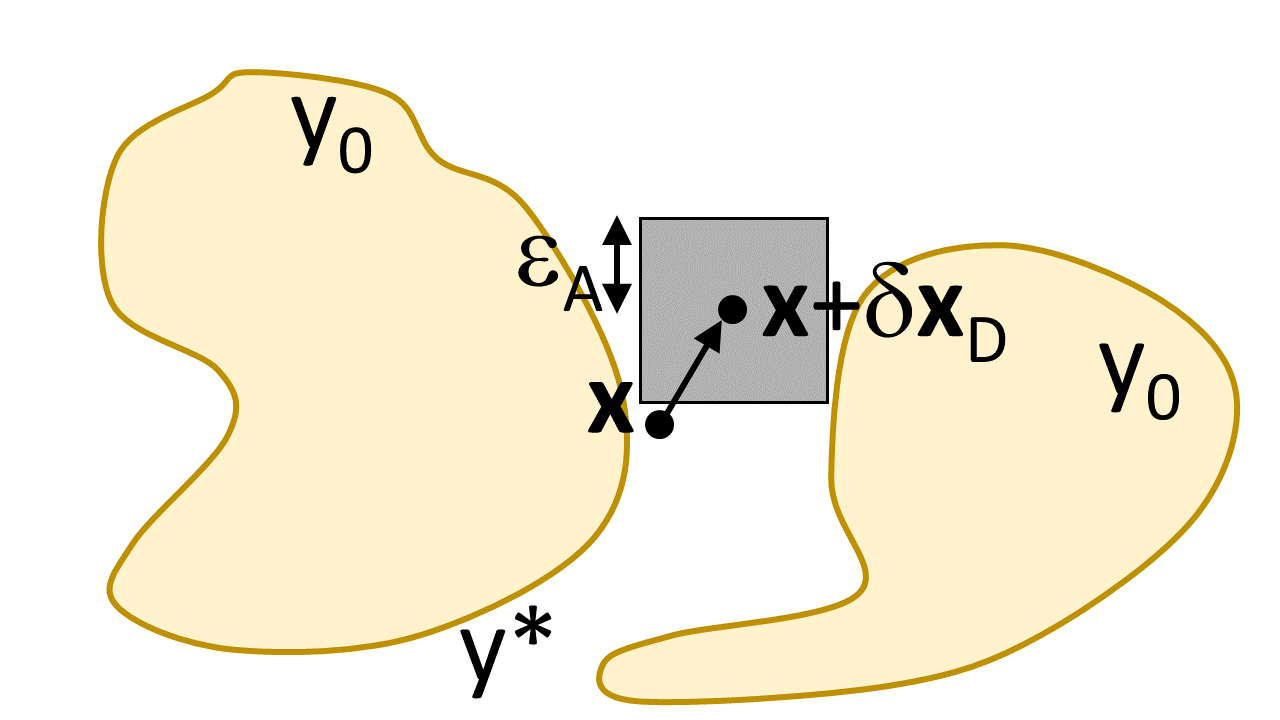} \\
(c) image purification & (d) $A^5$
\vspace{0.2cm} \\
\hline
\end{tabular}
\vspace{0.3cm}
\includegraphics[width=0.4\textwidth, trim=6cm 14.8cm 23.5cm 
2.7cm, clip]{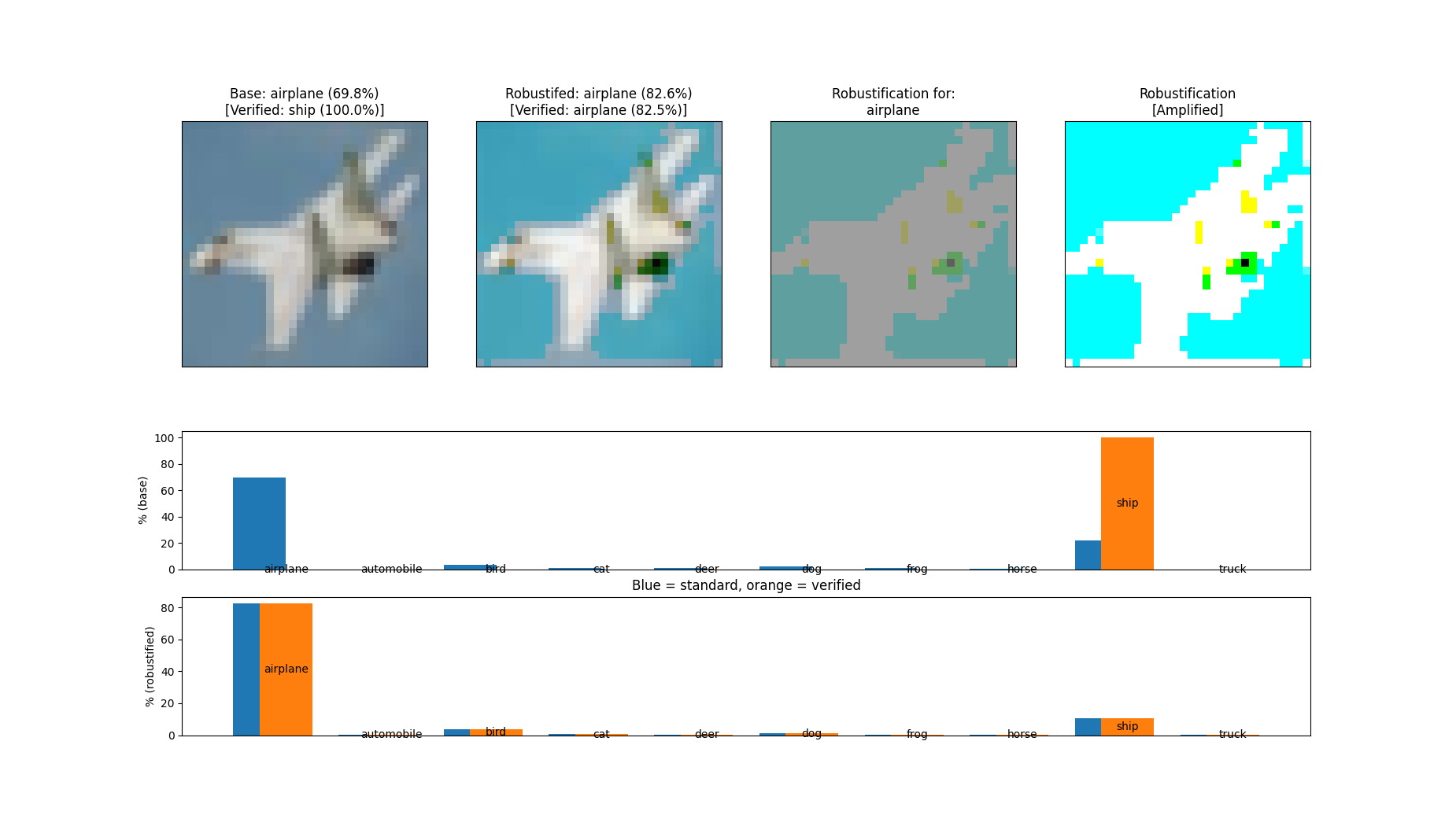}
\vspace{-0.3cm}
\caption{(a) An input $\boldsymbol{x} \in y*$ is correctly classified. The grey box (certified bounds~\cite{ibp_Gow19,crown_Zha18,crownibp_Zha20}) shows that, under an attack $\delta\boldsymbol{x}_A$, $||\delta\boldsymbol{x}_A||_\infty < \epsilon_A$, misclassification ($(\boldsymbol{x}+\delta\boldsymbol{x}_A) \in y_0$) is possible.
(b) Adversarial training~\cite{advtra_Bai21,advtra_Pan21} creates robust DNNs with regular classification landscapes: misclassification is less likely.
(c) Image purification~\cite{pure_Yoo21,onlinepur_Shi21} moves the attacked input $\boldsymbol{x}+\delta\boldsymbol{x}_A$ by $\delta\boldsymbol{x}_P$ back to correct classification. (d)
$A^5$ preemptively moves $\boldsymbol{x}$ into a non attackable position.
The original CIFAR10 image classified as \emph{airplane} ($69.8\%$ confidence) can be misclassified under attack (\emph{ship}, $100\%$ confidence). Once robustified through $A^5$ (right), misclassification does not occur anymore.}
\label{fig:teaser}
\end{figure}

Since Deep Neural Networks (DNNs) have been found vulnerable to adversarial attacks~\cite{intriguing_Sze14, adversarial_Goo15}, researchers studied various protection  strategies~\cite{advtra_Pan21,advtra_Bai21,crown_Zha18,crownibp_Zha20,ibp_Gow19}.
For instance, \emph{adversarial training}~\cite{intriguing_Sze14, adversarial_Goo15} generates attacks while asking a DNN for the correct output in training; it is simple, partially effective and widely adopted.
Certified methods (\eg, IBP~\cite{ibp_Gow19}, CROWN~\cite{crown_Zha18}, CROWN-IBP~\cite{crownibp_Zha20}) do a step more by estimating correct (although often pessimistic) output bounds (Fig.~\ref{fig:teaser}, a) used for training.
Adversarial training regularizes the classification landscape against the attacks (Fig.~\ref{fig:teaser}, b), but high protection often produces a loss in clean accuracy.
Other partially effective defenses are based on  randomness~\cite{randommitigation_Xie18,randommitigation_Xie18} or removal of the adversarial signal~\cite{pure_Yoo21,onlinepur_Shi21, Pure_Nie22,Son21_preprocessingaudio}, by moving the input back to the space of natural (non-attacked) data before classification (Fig.~\ref{fig:teaser}, c).

%\label{sec:notation}

\begin{table*}[!htp]
\centering
\resizebox{\textwidth}{!}{
\begin{tabular}{rlrl}
%\multicolumn{4}{c}{\bf{Notation and algorithms}}\\
\toprule
%$\boldsymbol{M}$ & A matrix &$\boldsymbol{v}$ & A vector\\
%$\boldsymbol{M}_j$ & $j$-th matrix row & $v_j$ & $j$-th vector element\\
$\boldsymbol{v}$ & A vector & $v_j$ & Vector $j$-th element\\
$C$ & Classifier &
$R$ & Robustifier\\
\midrule
$\boldsymbol{x}$ & Data &
$\delta\boldsymbol{x}_A$ & Attacking perturbation\\
$\delta\boldsymbol{x}_D$ & Defensive perturbation &
$\boldsymbol{x}+\delta\boldsymbol{x}_A$ & Data under attack\\
$\boldsymbol{x}+\delta\boldsymbol{x}_D$ & Robustified data &
$\boldsymbol{x}+\delta\boldsymbol{x}_D+\delta\boldsymbol{x}_A$ & Robustified data under attack\\
$\epsilon_D$ & Defense magnitude, $||\delta\boldsymbol{x}_D||_p<\epsilon_D$ &
$\epsilon_A^R$ & Attack magnitude, $||\delta\boldsymbol{x}_A||_p<\epsilon_A^R$ while training $R$\\
$\epsilon_A$& Attack magnitude $||\delta\boldsymbol{x}_A||_p<\epsilon_A$ while testing &
$\epsilon_A^C$ & Attack magnitude, $||\delta\boldsymbol{x}_A||_p<\epsilon_A^C$ while training $C$\\
%\midrule
%$A^5/O$ & \multicolumn{3}{l}{Offline robustification with known ground truth, legacy classifier $C$.} \\
%$A^5/O$ & \multicolumn{3}{l}{Robustify inputs for defense against white/black box, MitM attacks, when the ground truth class is known. Use the legacy classifier $C$.} \\
%$A^5/R$ & \multicolumn{3}{l}{On-the-fly robustification with a trained robustifier $R$, legacy classifier $C$, unknown ground truth.} \\
%$A^5/R$ & \multicolumn{3}{l}{Train a robustifier $R$ for on-the-fly defense against white/black box, MitM attacks, when the ground truth class is unknown. Use the legacy classifier $C$.} \\
%$A^5/RC$ & \multicolumn{3}{l}{On-the-fly robustification with a trained robustifier $R$, trained classifier $C$, unknown ground truth.} \\
%$A^5/RC$ & \multicolumn{3}{l}{Train a robustifier $R$ for on-the-fly defense against white/black box, MitM attacks, when the ground truth class is unknown. Retrain the classifier $C$.} \\
%$A^5/P$ & \multicolumn{3}{l}{Create robust prototypes $P$ for defense against white/black box, MitM/physical$^*$ attacks. Use the legacy classifier $C$.} \\
%$A^5/PC$ & \multicolumn{3}{l}{Offline robustification of prototypes $P$, trained classifier $C$, known ground truth.} \\
%$A^5/PC$ & \multicolumn{3}{l}{Create robust prototypes $P$ for defense against white/black box, MitM/physical$^*$ attacks. Retrain the classifier $C$.} \\
\bottomrule
%\multicolumn{4}{l}{$^*$ \small{Protection against physical attacks is not tested here.}}
\end{tabular}}
\caption{Notation for data $\boldsymbol{x}$, that are the inputs of the classifier $C$. We adopt an equivalent notation for physical objects $\boldsymbol{w}$.}
\label{tab:notation}
\end{table*}

All the aforementioned strategies activate the defense mechanism only \emph{after} the attack has been crafted. 
However, when dealing with adversarial attacks, \emph{the first actor to move has a significant advantage}.
For instance, perturbing an image can avoid online person identification and preserve privacy~\cite{privacy_Che21, privacy_Oh17, privacy_Sha20}.
Acting first is particularly suitable against Man in the Middle (MitM) attacks that may practically arise in  automotive~\cite{Wan21_maninthemiddle}, audio processing~\cite{Qin19_audiomitm, Yak19_audiophys, Son21_preprocessingaudio}, or while communicating with a remote machine~\cite{Yan22_RobustSense}.
Our idea spouses this reasoning line: we investigate a novel way to augment the data to preemptively \emph{certify} that it cannot be attacked (Fig.~\ref{fig:teaser}, d).
This fits in a recent research area that up to now has received less attention than adversarial training.
Researchers explored image and real object augmentation to guarantee a high recognition rate in presence of noise or geometric distortions, but not in adversarial scenarios~\cite{Unadversarial_Sal21}. 
Encryption schemes coupled with deep learning~\cite{Apr20_encrypt, Shi21__homoencr} or watermarking \cite{Li21_watermarking} only partially protect against MitM attacks.
The few existing preemptive robustification methods~\cite{Rak18_blind} include a procedure~\cite{preemptive_Moo22} that first runs a classifier on clean data and achieves preemptive robustification against MitM attacks through iterative optimization; these are partially effective and do not provide any certification.
Our novel framework encompasses most of the aforementioned cases while also introducing for the first time the concept of certified robustification.
More in detail, our manifold contributions are:
\begin{enumerate}[label=(\roman*), parsep=0pt, itemsep=0pt]
    \item We introduce $A^5$ (Adversarial Augmentation Against Adversarial Attacks), a comprehensive framework for preemptive augmentation to make data and physical objects \emph{certifiably} robust against MitM adversarial attacks. As far as we know, this is the first time certified defense is achieved in this way. Since we provide certified robustness, we guarantee protection against any form of white, grey or black box attack.
    \item We test different flavours of $A^5$ on standard datasets. By doing so, we study the connection between the robustness of the legacy classifier, the magnitude of the defensive augmentation, and the protection level delivered by $A^5$. We show $A^5$ achieving state-of-the-art certified protection against MitM attacks by training a robustifier DNN coupled with the legacy classifier, and even better results for co-training of the robustifier and classifier. We perform a critical, visual inspection of the robustified images to answer an interesting theoretical question (\emph{how does a non-attackable image look like?}) and potentially provide directions for the acquisition of inherently robust images.
    \item Using Optical Character Recognition (OCR) as an example, we show the application of $A^5$ for the design of certifiably robust physical objects, which extends~\cite{Unadversarial_Sal21} to the case of certified defense against MitM attacks.
    \item We share our code at \url{https://github.com/NVlabs/A5}, to allow replicating our results or testing $A^5$ in scenarios not considered here, for instance on other datasets or for protection against physical adversarial attack (\eg, adversarial patches).
\end{enumerate}

\FloatBarrier
\section{Related Work and Background}
\label{sec:related_work}

We identify three main philosophies for the development of defenses against adversarial attacks.
They are not mutually exclusive and can be adopted together.

The first approach aims at creating DNNs that are robust to adversarial attacks: \eg, in adversarial training, attacks are generated while training the DNN that should process them correctly~\cite{adversarial_Goo15,advtra_Pan21,advtra_Bai21}.
This regularizes the classification landscape and makes  attacks less effective (Fig.~\ref{fig:teaser}, b).
Adversarial training is simple but has one drawback: as attack generation is computationally demanding (even after speed up~\cite{Sha19_free}), attacks are randomly sampled and cannot cover the entire input space, thus impacting the final accuracy.
Certified defense~\cite{crown_Zha18, ibp_Gow19, crownibp_Zha20} can be seen as a state of the art 
refined form of adversarial training: robustness is obtained by propagating through the DNN an input interval that encompass all the attacks with magnitude up to $\epsilon_A^C$.
The corresponding set of output bounds allows estimating the worst case scenario and derive the training cost function (\eg, worst case entropy~\cite{ibp_Gow19, crownibp_Zha20}).
Research focuses on finding bounds that are both tight and computationally light.
IBP~\cite{ibp_Gow19} computes (and backwards) the bounds at the cost of an additional forward and backward pass. Since the bounds are loose at the beginning of training, a complex schedule is required to guarantee stability and convergence. Linear relaxation~\cite{crown_Zha18} delivers tighter bounds leveraging the correlation between different layers to linearly propagate signal and bounds, when possible, and eventually put in linear relation the input interval and output bounds.
CROWN~\cite{crown_Zha18} bounds are often tighter than IBP, but at an impractical computational cost for large DNNs.
Among the attempts to speed up CROWN~\cite{autolirpa_Xu21,betacrown_Wang21}, CROWN-IBP~\cite{crownibp_Zha20} strikes the right balance of computational complexity and tightness.
It is implemented in the auto{\_}LiRPA\footnote{Automatic Linear Relaxation based Perturbation Analysis} library~\cite{lirpa_Xu20, autolirpa_Xu21} adopted here. 
One limitation of adversarial training, certified methods, and $A^5$ as well, is the trade-off between clean and certified (worst case under-attack) accuracy: increasing the latter leads to a decrease in the former~\cite{advtra_Bai21,advtra_Pan21,ibp_Gow19,crownibp_Zha20}. 
Among the architecture solutions against adversarial attacks, we also mention the recent development of $\ell_\infty$-dist neurons that are inherently robust~\cite{Zha21_LInf}, and could be leveraged by $A^5$ as well in future.

The second defense philosophy is based on active countermeasures taken \emph{after} the attack is crafted. Image purification~\cite{pure_Yoo21,onlinepur_Shi21, Pure_Nie22} projects an attacked image onto the manifold of natural images through a generative model (Fig.~\ref{fig:teaser}, c).
It is less effective than adversarial training~\cite{verify_Cro20}, but does not make any assumption on the classifier.
Other forms of signal pre-processing have been proposed in the field of audio processing and automatic speech recognition, where MitM physical attacks can be easily implemented~\cite{Qin19_audiomitm, Yak19_audiophys} and partially defeated by randomized smoothing,
DefenseGAN, variational autoencoder (VAE), and Parallel
WaveGAN vocoder~\cite{Son21_preprocessingaudio}.
Both certified defenses and pre-processing methods do not assume any specific attack form, but the latter are less effective in white box scenarios~\cite{obfuscation_Ath18,adaptive_Tra20}.
Randomization can also be used after the attack, \eg by  changing resolution or padding an image~\cite{randommitigation_Xie18}, or by picking random experts in a mixture, that is proved to be more robust than deterministic DNNs~\cite{randommatters-Pin20}.

$A^5$ belongs to a third class of methods, that put in place countermeasures \emph{before} the attack (Fig.~\ref{fig:teaser}, d).
The roots of $A^5$ lie in two ideas that have been only partially investigated so far.
The first is that, in adversarial scenarios, \emph{the first actor to move has a significant advantage}.
This rule is already leveraged for online privacy, where a preemptive adversarial attack prevents the identification of the framed subject~\cite{privacy_Che21,privacy_Oh17}.
Since adversarial attacks often generalize to many classifiers, this ploy is particularly effective~\cite{privacy_Sha20}.
Preemptive protection is also a natural form of defense against MitM attacks that can be easily created in  contexts like automotive~\cite{Wan21_maninthemiddle}, speech and audio processing~\cite{Qin19_audiomitm, Yak19_audiophys, Son21_preprocessingaudio}, or anytime the capturing device communicates with a server~\cite{Yan22_RobustSense}.
Encryption~\cite{Apr20_encrypt, Shi21__homoencr} and watermarking~\cite{Li21_watermarking} coupled with deep learning work against MitM attacks, but do not achieve full protection, are not certified, require the key to be hidden (grey box scenario) and may significantly alter the appearance of the image.
Blind-preprocessing~\cite{Rak18_blind} applies a tanh transform, normalization, quantization and thermometer encoding to achieve partial and  uncertified defense.
The closest algorithm to $A^5$ is preemptive robustification~\cite{preemptive_Moo22}: given an image $\boldsymbol{x}$ and a classifier $C$ that can be fine-tuned, it computes $y_0=C(\boldsymbol{x})$ and randomly samples attacks $\delta\boldsymbol{x}_A$ against $\boldsymbol{x}+\delta\boldsymbol{x}_D$ to find a defensive perturbation $\delta\boldsymbol{x}_D$, such that $(\boldsymbol{x}+\delta\boldsymbol{x}_D+\delta\boldsymbol{x}_A) \in y_0$ (correct classification under attack).
Since $A^5$ uses certified bounds, it results in simpler training and higher protection, and does not require running $C$ first, as our robustifier DNN generalizes to all the inputs.
The second idea behind $A^5$ comes from physical adversarial examples, \ie, 3d physical objects~\cite{RobustAdv_Ath18} or 2d patches~\cite{AdvPAtch_Aba17} that consistently fool a DNN under a wide range of viewing conditions.
Reverting the attacker and defender roles, the complementary cost function leads to the creation of \emph{unadversarial examples}: patches or textures that, once applied to a real 3d object, reinforce (instead of degrading) the DNN desired behaviour~\cite{Unadversarial_Sal21}. However, these are not designed to be inherently robust to adversarial attacks.
We exploit the idea in $A^5$ in all those situations where system designers control, to some extent, the inputs fed to the DNN (\eg in OCR, robotic systems or during the design of the road infrastructure).
We show as an example the use of $A^5$ to design \emph{certifiably protected} characters for OCR, something unexplored both in~\cite{preemptive_Moo22} and~\cite{Unadversarial_Sal21}.

\begin{figure*}[th!]
\centering
\includegraphics[width=\textwidth,clip=True, trim=0cm 0cm 2.5cm 0cm]{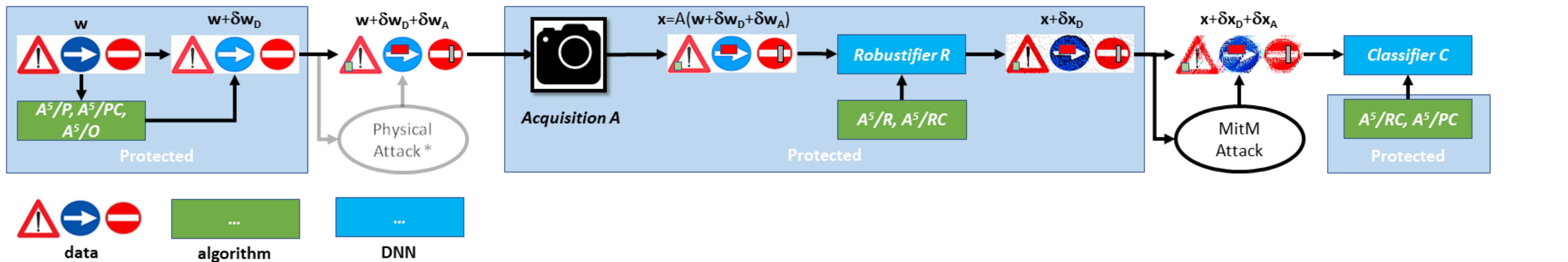} \caption{Schematic representation of the full $A^5$ framework. It allows the robust augmentation of both physical objects $\boldsymbol{w}$ and acquired data $\boldsymbol{x}$ to make them certifiably robust against MitM and physical (not tested here) adversarial attacks.}
\label{fig:A5}
\end{figure*}

\section{Method}
\label{sec:method}

\paragraph{Threat models} We summarize our notation in Table~\ref{tab:notation}, whereas Fig.~\ref{fig:A5} illustrates the full $A^5$ framework including trainable DNNs in blue, training algorithms in green,  physical objects $\boldsymbol{w}$, and acquired data $\boldsymbol{x}$.
MitM attacks are crafted while transmitting data from the acquisition device to the classifier $C$, while physical attacks (not tested here) are crafted onto physical objects $\boldsymbol{w}$, before data acquisition.
$A^5$ can be used in a scenario where the attacker has full access to the $C$: since it uses certified bounds, it is agnostic to the specific nature of the attack.
In other words, we certify protection for any $\delta \boldsymbol{x}_A$, $||\delta \boldsymbol{x}_A|| < \epsilon_A$ and the attacker can use any white-, grey- or black-box attacking algorithm.
We assume that $A^5$ can run in a protected environment, not accessible to the attacker, to perform robustification (light blue rectangles in Fig.~\ref{fig:A5}).
More details and limitations of our threat model are discussed in the Appendix, whereas here we first introduce useful definitions and then provide several recipes for the use of $A^5$ in different scenarios.

\paragraph{Definitions} For a classifier $C$ and input $\boldsymbol{x}$, $\boldsymbol{y} = C(\boldsymbol{x})$ is the logit vector output by $C$. Under attack, we have:
\begin{equation}
\boldsymbol{y} + \delta\boldsymbol{y} = C(\boldsymbol{x} + \delta\boldsymbol{x}_A), \; ||\delta\boldsymbol{x}_A||_p < \epsilon_A,
\label{eq:attack}
\end{equation}
where $\epsilon_A$ is the magnitude of the attack $\delta\boldsymbol{x}_A$ in $p$-norm and $\delta\boldsymbol{y}$ is the logit perturbation.
Linear relaxation~\cite{crown_Zha18} estimates the lower ($\boldsymbol{y}^l$) and upper ($\boldsymbol{y}^u$) bounds of $\boldsymbol{y} + \delta\boldsymbol{y}$, for instance using IBP~\cite{ibp_Gow19}, CROWN~\cite{crown_Zha18}, CROWN-IBP~\cite{crownibp_Zha20}, or a combination of them.
The margin for the class $j$ is the difference between the lower bound of the ground truth class $j^*$ and the upper bound of $j$, while it is zero by definition for $j^*$. More formally, for $M$ classes the elements of the margin vector $\boldsymbol{m}=[m_0, m_1, ..., m_{M-1}]$ are: $m_{j^*}=0 $ and $m_j=\boldsymbol{y}^l_{j^*}- \boldsymbol{y}^u_j,~\forall{j\neq j^*}$. 
The confidence is obtained by applying softmax to $\boldsymbol{m}$:
\begin{equation}
p_j(\boldsymbol{x}) = {e^{-m_j}}/{\sum_{k=0}^{M-1}e^{-m_k}},
\label{eq:worst_case_prob}
\end{equation}
and thus the worst case cross entropy $E(\boldsymbol{x})$ is:
\begin{equation}
E(\boldsymbol{x}) = -\sum_{j=0}^{M-1}{\hat{y}_j log(p_j(\boldsymbol{x}))},
\label{eq:single_image_cross_entropy}
\end{equation}
where $\hat{y}_j$ is the one-hot-encoding of $j^*$.

To formally introduce $A^5$, we first give the following definitions:
\begin{defn}[$\epsilon_A$-robustness] Given an input $\boldsymbol{x}$ and a classifier $C$, we say that $\boldsymbol{x}$ is \emph{$\epsilon_A$-robust with respect to $C$ in norm $p$} if $\forall {\delta\boldsymbol{x}_A}$, $||\delta\boldsymbol{x}_A||_p<\epsilon_A$,  $C(\boldsymbol{x}+\delta\boldsymbol{x}_A)$ generates the correct classification output.
\end{defn}
\begin{defn}[robustification] For a given input $\boldsymbol{x}$ and classifier $C$, \emph{robustification} is the process of finding a defensive augmentation $\delta\boldsymbol{x}_D$, $||\delta\boldsymbol{x}_D||_p<\epsilon_D$, such that $\boldsymbol{x}+\delta\boldsymbol{x}_D$ is $\epsilon_A$-robust with respect to $C$ in norm $p$.
\end{defn}

We compute certified CROWN-IBP bounds through auto{\_}LiRPA~\cite{autolirpa_Xu21} and optimize $E(\boldsymbol{x} + \delta\boldsymbol{x}_D)$ to find preemptive, defensive perturbations $\delta\boldsymbol{x}_D$ that make $\boldsymbol{x}+\delta\boldsymbol{x}_D$ $\epsilon_A$-robust, following the training recipes provided here.
We also show how to find defensive perturbations $\delta\boldsymbol{w}_D$ for physical objects $\boldsymbol{w}$ using the same approach.

\paragraph{Offline robustification with ground truth ($A^5/O$):}
Consider a legacy classifier $C$ that has been certifiably trained (\eg through CROWN-IBP~\cite{crownibp_Zha20}) against attacks of magnitude $\epsilon_A^C$.
Given an input $\boldsymbol{x}$ and its class $j^*$, we want to find a defensive perturbation $\delta\boldsymbol{x}_D$, $||\delta\boldsymbol{x}_D||_p<\epsilon_D$, such that $\boldsymbol{x}+\delta\boldsymbol{x}_D$ is $\epsilon_A^R$-robust with respect to $C$ in $p$-norm.
Our solution to this problem is $A^5/\text{Offline}$ (or $A^5/O$).

To force $||\delta\boldsymbol{x}_D||_p<\epsilon_D$, we parameterize $\delta\boldsymbol{x}_D$ via a vector $\boldsymbol{z}$ to which we apply an element-wise sigmoid:
\begin{eqnarray}
\delta\boldsymbol{x}_D = \delta\boldsymbol{x}_D(\boldsymbol{z}) = [\delta x_D(\boldsymbol{z})_0, \dots, \delta x_D(\boldsymbol{z})_{K-1}]\\
\delta x_D(\boldsymbol{z})_j = 2 \epsilon_D [1/(1+e^{-z_j})-0.5],%\alpha z_j})-0.5],
\label{eq:norm_constraint}
\end{eqnarray}
where $K$ is the number of elements in $\boldsymbol{x}$. 
This is a robust classification problem where $\boldsymbol{z}$ is the only unknown. %, and for now we set $\alpha=1$.
For any $\boldsymbol{x}$, we minimize the worst case cross entropy:
\begin{equation}
\boldsymbol{z} = \argmin_{\tilde{\boldsymbol{z}}} E[\boldsymbol{x}+\delta\boldsymbol{x}_D(\tilde{\boldsymbol{z}})],
\label{eq:cost_function_a5o}
\end{equation}
using RMSProp.
Notice that the magnitude of the attack used during robustification ($\epsilon_A^R$) and that adopted to train $C$ ($\epsilon_A^C$) can be different. Furthermore, similarly to adversarial training and certified methods, these can also be different from the target attack magnitude, $\epsilon_A$.

$A^5/O$ is similar in spirit to preemptive image robustification~\cite{preemptive_Moo22}, but does not need $C$ to run on the clean input before robustification.
Although $A^5/O$ could be used in a similar way to robustify $\boldsymbol{x}$ accordingly to the class $C(\boldsymbol{x})$, running $C$ first and then iteratively solving the optimization problem in Eq.~\ref{eq:cost_function_a5o} makes it hardly usable in practice.
Therefore, it does not directly serve any practical purpose, but it helps answering important theoretical questions (\eg, under which conditions can we robustify $\boldsymbol{x}$? What does a robustified image looks like?) and establishing a baseline with known ground truth and no computational constraints.

\paragraph{Online robustification with robustifier $R$, legacy $C$ ($A^5/R$):}
We introduce a second recipe, $A^5/R$, where we train a robustifier DNN $R$ that takes in input $\boldsymbol{x}$, no ground truth label, and outputs $\boldsymbol{z}$, from which a defensive perturbation $\delta\boldsymbol{x}_D$, $||\delta\boldsymbol{x}_D||_p<\epsilon_D$, is computed as in Eq.~\ref{eq:norm_constraint}.
We assume again that a robust legacy $C$ is given and we train $R$ to make each $\boldsymbol{x}+\delta\boldsymbol{x}_D$ in the training dataset $\epsilon_A^R$-robust with respect to $C$ in $p$-norm.
%The magnitude of the attack used to train the classifier ($\epsilon_A^C$) and that used for the robustifier ($\epsilon_A^R$) can be different.
More formally, the weights $\boldsymbol{\theta}_R$ of $R$ are found by minimization of the average worst case entropy on the training dataset $E(\boldsymbol{\theta}_R)$, defined as:
\begin{eqnarray}
\boldsymbol{z} = R(\boldsymbol{x}|\boldsymbol{\theta}_R), ~
E(\boldsymbol{\theta}_R) = (1/N) \sum_{\boldsymbol{x}}{E[\boldsymbol{x}+\delta\boldsymbol{x}_D(\boldsymbol{z})]},
\end{eqnarray}
where $N$ is the dataset size. Unlike preemptive robustification~\cite{preemptive_Moo22}, $R$ does not iteratively solve a complex optimization problem, neither we need running $C$ before $R$.
Therefore, we can use $A^5/R$ to protect $\boldsymbol{x}$ against MitM adversarial attacks on-the-fly, ignoring its class, soon after its acquisition, and before its transmission to $C$ (see $A^5/R$ in Appendix).
Furthermore, since $R$ is trained for a specific $C$, it can be deployed on the field (\eg, into the firmware of the acquisition devices) without changing (and thus maintaining the legacy of) the classifier infrastructure.

\paragraph{Online robustification with robustifier $R$, retrained $C$  ($A^5/RC$):}
The $A^5/RC$ recipe is equivalent to $A^5/R$, but it leverages the co-adaptation of $C$ and $R$ by joint training (see $A^5/RC$ in Appendix). The cost function is again the average worst case entropy $E(\boldsymbol{\theta}_C, \boldsymbol{\theta}_R)$, defined now as:
\begin{eqnarray}
\boldsymbol{z} = R(\boldsymbol{x}|\boldsymbol{\theta}_R), \quad
\boldsymbol{y} = C(\boldsymbol{x}+\delta\boldsymbol{x}_D|\boldsymbol{\theta}_C), \quad \\
E(\boldsymbol{\theta}_C, \boldsymbol{\theta}_R) = (1/N) \sum_{\boldsymbol{x}}
E[\boldsymbol{x}+\delta\boldsymbol{x}_D(\boldsymbol{z})],
\end{eqnarray}
where $\boldsymbol{\theta}_C$ are the parameters of $C$.
Training $R$ and $C$ from scratch is often unstable and requires careful tuning to achieve convergence (likewise robust classifiers~\cite{ibp_Gow19,crownibp_Zha20}).
Therefore, we first train $R$ with $A^5/C$ and then fine tune $R$ and $C$ together.
During the tuning phase, we can use random input augmentation to prevent overfitting (see Section~\ref{sec:results} and Appendix for details).
Differently from $A^5/R$, $A^5/RC$ does not preserve the legacy classifier $C$, but it achieves the same scope (with more effectiveness) in terms of protection against MitM attacks.

\paragraph{Offline physical robustification with ($A^5/PC$) and without ($A^5/P$) retraining $C$, with ground truth:}
This recipe aims at producing physical objects that are certifiably robust against MitM  attacks (see $A^5/P$,  $A^5/PC$ in Appendix).
Similarly to unadversarial objects~\cite{Unadversarial_Sal21}, we assume that the system creator can design the entire infrastructure and the attacker cannot interfere in this phase.
The problem is formally defined for $A^5/PC$ as:
\begin{eqnarray}
\boldsymbol{x} + \delta\boldsymbol{x}_D = A(\boldsymbol{w}+\delta\boldsymbol{w}_D),~||\delta\boldsymbol{w}_D)||<\epsilon_D, \\
E(\boldsymbol{\theta}_C, \delta\boldsymbol{w}_D) = (1/N) \sum_{\boldsymbol{x}} E(\boldsymbol{x}+\delta\boldsymbol{x}_D)
\end{eqnarray}
where $\boldsymbol{w}$ is the object appearance, whereas $\boldsymbol{x}=A(\boldsymbol{w})$ is the data acquisition model (\eg, image formation).
$N$ is here the number of random samples generated as $\boldsymbol{x} + \delta\boldsymbol{x}_D  = A(\boldsymbol{w}+\delta\boldsymbol{w}_D)$; it is important noticing in fact that the acquisition process is not deterministic: as an exemplary case we mention OCR, where $\boldsymbol{w}$ is the shape of a character, while $A$ may include random perspective transformations, contrast and brightness changes, noise addition and blurring, that may occur on a real camera.
The defensive perturbation $\delta\boldsymbol{w}_D$ represents a change in the physical character appearance $\boldsymbol{w}$ that decreases the success rate of an attack towards the acquired image $\boldsymbol{x}$ (therefore after framing the character with $A$) and the classifier $C$.

In $A^5/P$ we change the character shapes and use the legacy classifier $C$;
 in $A^5/PC$ we also train $C$ while estimating the robust characters $\boldsymbol{w}+\delta\boldsymbol{w}$.
In both cases $C$ is previously trained to be robust, \eg, through CROWN-IBP~\cite{crownibp_Zha20}. 
The $A^5/P$ and $A^5/PC$ recipes find application in all those situations where the developer can design the entire infrastructure; beyond OCR, other examples are road sign or robotic tool design and classification.
\section{Results and discussion}
\label{sec:results}

We test $A^5/O$, $A^5/R$, and $A^5/RC$ on MNIST~\cite{mnist_Den12}, CIFAR10~\cite{cifar10_Kri09}, FashionMNIST~\cite{Xia17_fashion}, and Tinyimageenet~\cite{le2015tiny}, that have been widely used to establish significant milestones in the  adversarial defense field.
Training schedules and DNN architectures are given in the Appendix.
In all cases we use a $p=\infty$ norm for the attack and defense.
Our intention is to show the improvement
achieved by $A^5$ over traditional, certified methods like CROWN-IBP.
The reader should anyway keep in mind that the direct comparison of $A^5$ and CROWN-IBP is partially unfair, as CROWN-IBP does not apply any preemptive defense.

\begin{table}
\center
\resizebox{\columnwidth}{!}{\begin{tabular}{c|ccccc}
$\epsilon_A$ & $\epsilon_A^C = 0.0$ & $\epsilon_A^C = 0.1$ & $\epsilon_A^C = 0.2$ & $\epsilon_A^C = 0.3$ & $\epsilon_A^C = 0.4$ \\
\midrule
0.1 & --- & --- & 1.17 [---, 2.24]~\cite{crownibp_Zha20} & --- & --- \\
0.1 & 0.81 [89.02, 100.00] & 0.92 [2.00, 5.41] & 1.32 [2.14, 2.60] & 1.44 [2.33, 2.63] & 2.42 [3.36, 3.41] \\
\midrule
0.3 &  --- & --- & --- & 1.82 [---, 7.02]~\cite{crownibp_Zha20} & --- \\
0.3 & 0.69 [100.00, 100.00] & 0.94 [98.41, 100.00] & 1.30 [99.62, 100.00] & 1.41 [7.06, 9.98] & 2.58 [7.03, 7.68] \\
\end{tabular}}
\caption{Error on clean data and (within brackets) under autoattack~\cite{autoattack_1, autoattack_2} and certified, for MNIST classifiers trained by us with CROWN-IBP~\cite{crownibp_Zha20}, under attack $\epsilon_A = \{0.1, 0.3\}$; we also show the metrics reported in~\cite{crownibp_Zha20}. A $100\%$ certified error rate identifies DNNs that are not protected against attacks.}
\label{tab:mnist_classifiers}
\end{table}

\begin{table}
\center
\resizebox{\columnwidth}{!}{\begin{tabular}{cc|ccccc}
$\epsilon_A^R = \epsilon_A$ & $\epsilon_D$ & $\epsilon_A^C = 0.0$ & $\epsilon_A^C = 0.1$ & $\epsilon_A^C = 0.2$ & $\epsilon_A^C = 0.3$ & $\epsilon_A^C = 0.4$ \\
\midrule
0.1 & 0.05 & 0.79 [87.09, 100.00] & 0.82 [1.51, 2.89] & 1.22 [1.64, 1.86] & 1.34 [1.76, 1.87] & 2.38 [2.91, 2.92] \\
0.1 & 0.10 & 0.79 [86.51, 100.00] & 0.72 [1.21, 1.92] & 1.15 [1.43, 1.49] & 1.34 [1.50, 1.58] & 2.41 [2.71, 2.74] \\
0.1 & 0.20 & 0.89 [86.40, 100.00] & 0.56 [0.78, 1.04] & 1.13 [1.18, 1.21] & 1.35 [1.37, 1.38] & 2.40 [2.55, 2.55] \\
0.1 & 0.30 & 0.93 [86.44, 100.00] & 0.51 [0.57, 0.70] & 1.14 [1.16, 1.17] & 1.34 [1.37, 1.37] & 2.38 [2.52, 2.52] \\
0.1 & 0.40 & 0.99 [85.93, 100.00] & 0.52 [0.53, 0.60] & 1.14 [1.16, 1.16] & 1.35 [1.38, 1.38] & 2.44 [2.51, 2.51] \\
\midrule
0.3 & 0.05 & 0.71 [100.00, 100.00] & 0.93 [97.18, 100.00] & 1.34 [99.17, 100.00] & 1.35 [5.16, 6.97] & 2.55 [6.00, 6.42] \\
0.3 & 0.10 & 0.72 [100.00, 100.00] & 0.88 [95.32, 100.00] & 1.36 [98.58, 100.00] & 1.33 [4.12, 5.19] & 2.49 [5.54, 5.85] \\
0.3 & 0.20 & 0.72 [100.00, 100.00] & 0.88 [90.62, 100.00] & 1.41 [96.21, 100.00] & 1.32 [3.21, 3.77] & 2.35 [5.02, 5.22] \\
0.3 & 0.30 & 0.73 [100.00, 100.00] & 0.89 [86.28, 100.00] & 1.41 [94.18, 100.00] & 1.31 [2.91, 3.40] & 2.32 [4.66, 4.81] \\
0.3 & 0.40 & 0.81 [100.00, 100.00] & 0.85 [83.49, 100.00] & 1.42 [92.63, 100.00] & 1.32 [2.74, 3.20] & 2.28 [4.16, 4.29] \\
\end{tabular}}
\caption{Error on clean data and (within brackets) under autoattack~\cite{autoattack_1, autoattack_2} and certified, for MNIST, $A^5/O$, under attack $\epsilon_A = \{0.1, 0.3\}$. During training we use $\epsilon_A^R = \epsilon_A$.}
\label{tab:mnist_a5o}
\end{table}

\begin{table}
\center
\resizebox{\columnwidth}{!}{\begin{tabular}{cc|ccccc}
$\epsilon_A^R = \epsilon_A$ & $\epsilon_D$ & $\epsilon_A^C = 0.0$ & $\epsilon_A^C = 0.1$ & $\epsilon_A^C = 0.2$ & $\epsilon_A^C = 0.3$ & $\epsilon_A^C = 0.4$ \\
\midrule
0.1 & 0.05 & 0.82 [85.50, 100.00] & 0.95 [1.71, 3.58] & 1.30 [1.82, 2.02] & 1.40 [1.86, 2.03] & 2.43 [2.96, 2.97] \\
0.1 & 0.10 & 0.81 [82.87, 100.00] & 1.00 [1.48, 2.54] & 1.19 [1.62, 1.73] & 1.40 [1.68, 1.74] & 2.38 [2.77, 2.79] \\
0.1 & 0.20 & 0.88 [78.48, 100.00] & 0.93 [1.33, 1.77] & 1.20 [1.42, 1.52] & 1.31 [1.53, 1.59] & 2.26 [2.55, 2.57] \\
0.1 & 0.30 & 0.99 [75.89, 100.00] & 0.91 [1.16, 1.54] & 1.14 [1.31, 1.36] & 1.26 [1.43, 1.47] & 2.20 [2.42, 2.43] \\
0.1 & 0.40 & 1.04 [75.69, 100.00] & 0.87 [1.05, 1.35] & 1.14 [1.27, 1.33] & 1.24 [1.43, 1.46] & 2.22 [2.41, 2.43] \\
\midrule
0.3 & 0.05 & 0.69 [100.00, 100.00] & 0.94 [96.93, 100.00] & 1.33 [99.11, 100.00] & 1.41 [5.13, 6.91] & 2.58 [5.77, 6.09] \\
0.3 & 0.10 & 0.71 [100.00, 100.00] & 0.91 [95.09, 100.00] & 1.33 [98.47, 100.00] & 1.38 [3.86, 4.81] & 2.55 [4.71, 4.83] \\
0.3 & 0.20 & 0.72 [100.00, 100.00] & 0.92 [90.13, 100.00] & 1.37 [96.08, 100.00] & 1.41 [2.26, 2.63] & 2.55 [3.33, 3.40] \\
0.3 & 0.30 & 0.75 [100.00, 100.00] & 0.93 [85.54, 100.00] & 1.35 [93.36, 100.00] & 1.32 [1.86, 1.99] & 2.37 [2.86, 2.88] \\
0.3 & 0.40 & 0.83 [100.00, 100.00] & 0.96 [82.48, 100.00] & 1.36 [92.15, 100.00] & 1.29 [1.69, 1.83] & 2.35 [2.67, 2.69] \\
\end{tabular}}
\caption{Error on clean data and (within brackets) under autoattack~\cite{autoattack_1, autoattack_2} and certified, for MNIST, $A^5/R$, under attack $\epsilon_A = \{0.1, 0.3\}$. During training we use $\epsilon_A^R = \epsilon_A$.}
\label{tab:mnist_a5r}
\end{table}

\begin{table}
\center
\resizebox{\columnwidth}{!}{\begin{tabular}{cc|ccccc}
$\epsilon_A^R = \epsilon_A$ & $\epsilon_D$ & $\epsilon_A^C = 0.0$ & $\epsilon_A^C = 0.1$ & $\epsilon_A^C = 0.2$ & $\epsilon_A^C = 0.3$ & $\epsilon_A^C = 0.4$ \\
\midrule
0.1 & 0.05 & 88.66 [88.66, 88.66] & 1.00 [1.38, 2.35] & 1.16 [1.61, 2.46] & 1.14 [1.54, 2.29] & 1.37 [1.72, 2.33] \\
0.1 & 0.10 & 88.66 [88.66, 88.66] & 0.77 [0.95, 1.40] & 0.99 [1.16, 1.56] & 0.86 [1.03, 1.42] & 1.13 [1.31, 1.72] \\
0.1 & 0.20 & 88.66 [88.66, 88.66] & 0.69 [0.80, 1.05] & 0.84 [0.94, 1.25] & 0.96 [1.07, 1.33] & 0.88 [1.10, 1.54] \\
0.1 & 0.30 & 88.66 [88.66, 88.66] & 0.64 [0.70, 1.02] & 0.91 [0.99, 1.16] & 0.79 [0.88, 1.21] & 1.00 [1.12, 1.43] \\
0.1 & 0.40 & 88.66 [88.66, 88.66] & 0.61 [0.68, 0.88] & 0.93 [0.99, 1.18] & 0.69 [0.76, 1.10] & 1.00 [1.03, 1.30] \\
\midrule
0.3 & 0.05 & 88.66 [88.66, 88.66] & 88.66 [88.66, 88.66] & 2.57 [8.35, 10.86] & 3.42 [9.51, 11.72] & 8.00 [18.05, 19.77] \\
0.3 & 0.10 & 88.66 [88.66, 88.66] & 88.66 [88.66, 88.66] & 2.07 [5.66, 7.56] & 2.35 [6.18, 8.07] & 2.35 [6.31, 7.93] \\
0.3 & 0.20 & 88.66 [88.66, 88.66] & 1.68 [3.11, 4.79] & 1.40 [2.42, 3.62] & 1.28 [2.82, 4.01] & 1.88 [3.53, 4.90] \\
0.3 & 0.30 & 88.66 [88.66, 88.66] & 0.84 [0.96, 1.38] & 1.00 [1.14, 1.36] & 0.94 [1.15, 1.59] & 1.07 [1.39, 1.90] \\
0.3 & 0.40 & 88.66 [88.66, 88.66] & 0.72 [0.80, 1.03] & 0.75 [0.82, 1.06] & 0.91 [1.14, 1.68] & 0.87 [1.14, 1.50] \\
\end{tabular}}
\caption{Error on clean data and (within brackets) under autoattack~\cite{autoattack_1, autoattack_2} and certified, for MNIST, $A^5/RC$, under attack $\epsilon_A = \{0.1, 0.3\}$. During training we use $\epsilon_A^R = \epsilon_A$, and data augmentation.}
\label{tab:mnist_a5rc_augment}
\end{table}

We first use CROWN-IBP to train five certifiably robust MNIST classifiers with varying levels of robustness, for training attacks $\epsilon_A^C=0.0$ (unprotected $C$) and $\epsilon_A^C=\{0.1, 0.2, 0.3, 0.4\}$.
Their clean and certified errors are reported in Table~\ref{tab:mnist_classifiers}, for attacks $\epsilon_A = \{0.1, 0.3\}$.
We also report the error rates estimated with autoattack~\cite{autoattack_1, autoattack_2}, an automatic toll based on an ensemble of four attacks to evaluate the practical DNN robustness.
These error rates are in the same ballpark of those reported in the CROWN-IBP paper~\cite{crownibp_Zha20}: they represent the baseline for the analysis of the different $A^5$ recipes considered here.

For each $C$ we run $A^5/O$ on the MNIST test set, for defense magnitudes $\epsilon_D=\{0.05, 0.1, 0.2, 0.3, 0.4\}$.
During the optimization of the entropy in Eq.~\ref{eq:cost_function_a5o}, we use attacks  $\epsilon_A^R=\{0.1, 0.3\}$, not necessarily equal to $\epsilon_A^C$.
Table~\ref{tab:mnist_a5o} reports the clean, autoattack, and certified errors for a testing attack $\epsilon_A=\epsilon_A^R$.
When $C$ is not robust ($\epsilon_A^C=0.0$), $A^5/O$ does not reduce the certified error\footnote{Coherently with~\cite{Unadversarial_Sal21}, the worst case cross entropy improves, but not enough to affect the ranking of the classified classes under attack.}, otherwise it generally overcomes the base CROWN-IBP classifier, as shown by comparing against the error rates in Table~\ref{tab:mnist_classifiers}. 
As a rule of thumb, the best clean / certified error trade off is often achieved for $\epsilon_A^R=\epsilon_A^C=\epsilon_A$.
Results also improve, as expected, for larger defenses $\epsilon_D$, when $A^5/O$ beat CROWN-IBP by a significant margin on clean and certified errors.
However, $A^5/O$ does not work well when the legacy $C$ is trained with a large $\epsilon_A^C$, possibly because of its initial low clean accuracy.
The results are consistent with our interpretation of the effects of adversarial training and robustification on the classification landscape (Fig.~\ref{fig:teaser}): training $C$ with a large $\epsilon_A^C$ creates classification valleys wide enough to accommodate $\boldsymbol{x}+\delta\boldsymbol{x}_D+\delta\boldsymbol{x}_A$, rendering the adversarial attacks ineffective.
$A^5$ moves any $\boldsymbol{x}$ towards the centers of the valleys; a large $\epsilon_D$ allows moving them more easily.
Too large $\epsilon_A^C$ remain however detrimental for the clean accuracy.

Table~\ref{tab:mnist_a5r} reports results for $A^5/R$, demonstrating that training a DNN robustifier $R$ for on-the-fly, certified robustification, ignoring the ground truth, is feasible. Likewise $A^5/O$, $A^5/R$  beats the base CROWN-IBP classifier by a significant margin. Sometimes $A^5/O$ does better by leveraging the knowledge of the ground truth class, whereas the regular robustification signal generated by $A^5/R$ is more effective in other cases.

Our last MNIST experiment investigates $A^5/RC$ that does not preserve the legacy $C$ like $A^5/R$, but achieves better clean accuracy and protection (Table~\ref{tab:mnist_a5rc_augment}).
In the Appendix we discuss the role of data augmentation to achieve this result.
Interestingly, the best results are obtained when $C$ is initially trained for $\epsilon_A^C\leq\epsilon_A$, probably because $C$ has an initial high clean accuracy and subsequently co-adapts with $R$ to also guarantee a small certified error. 

\begin{table}
\center
\resizebox{\columnwidth}{!}{\begin{tabular}{cc|ccc}
$\epsilon_A^C$ & $\epsilon_D$ & $\epsilon_A^R = 4/255$ & $\epsilon_A^R = 8/255$ & $\epsilon_A^R = 16/255$ \\
\midrule
8/255 & 4/255 & 52.75 [63.47, 64.09] & 53.34 [62.42, 62.79] & 53.75 [64.92, 65.49] \\
8/255 & 8/255 & 51.06 [60.04, 60.73] & 52.36 [59.24, 59.58] & 53.16 [63.50, 64.03] \\
8/255 & 16/255 & 48.72 [54.89, 55.25] & 50.60 [55.22, 55.38] & 52.23 [60.79, 61.14] \\
8/255 & 32/255 & 45.68 [49.51, 49.74] & 48.06 [50.86, 50.94] & 50.99 [57.37, 57.61] \\
\end{tabular}}
\caption{Error on clean data and (within brackets) under autoattack~\cite{autoattack_1, autoattack_2} and certified, for CIFAR10, $A^5/O$, under attack $\epsilon_A = 8 / 255$. During training we use different $\epsilon_A^R$ and $\epsilon_D$.
The robust classifier trained with CROWN-IBP has clean [and certified] errors equal to 54.02 [66.94] or 45.47 [69.55], depending on the CROWN-IBP training configuration~\cite{crownibp_Zha20}.}
\label{tab:cifar10_a5o}
\end{table}

\begin{table}
\center
\resizebox{\columnwidth}{!}{\begin{tabular}{cc|ccc}
$\epsilon_A^C$ & $\epsilon_D$ & $\epsilon_A^R = 4/255$ & $\epsilon_A^R = 8/255$ & $\epsilon_A^R = 16/255$ \\
\midrule
8/255 & 0/255 & 54.60 [66.34, 67.11] & 54.60 [66.34, 67.11] & 54.60 [66.31, 67.11] \\
\midrule
8/255 & 4/255 & 53.73 [64.07, 64.57] & 54.12 [63.32, 63.74] & 54.37 [63.58, 64.03] \\
8/255 & 8/255 & 52.91 [62.59, 63.12] & 53.24 [61.38, 61.93] & 54.54 [60.85, 61.16] \\
8/255 & 16/255 & 51.58 [60.40, 61.02] & 51.95 [59.21, 59.67] & 54.44 [59.87, 60.11] \\
8/255 & 32/255 & 50.91 [57.26, 57.75] & 51.70 [56.91, 57.43] & 54.05 [58.39, 58.73] \\
\end{tabular}}
\caption{Error on clean data and (within brackets) under autoattack~\cite{autoattack_1, autoattack_2} and certified, for CIFAR10, $A^5/R$, under attack $\epsilon_A = 8 / 255$. During training we use different $\epsilon_A^R$ and $\epsilon_D$.
The robustifier $R$ in the first row uses $\epsilon_D = 0$ and thus it does not provide any form of defense --- the error reported here is equivalent to that of a  robust classifier trained with CROWN-IBP~\cite{crownibp_Zha20}.}
\label{tab:cifar10_a5r}
\end{table}

\begin{table}
\center
\resizebox{\columnwidth}{!}{\begin{tabular}{cc|ccc}
$\epsilon_A^C$ & $\epsilon_D$ & $\epsilon_A^R = 4/255$ & $\epsilon_A^R = 8/255$ & $\epsilon_A^R = 16/255$ \\
\midrule
8/255 & 0/255 & 50.24 [69.17, 73.65] & 58.63 [69.25, 70.63] & 74.28 [77.82, 77.91] \\
\midrule
8/255 & 4/255 & 46.76 [63.03, 70.99] & 54.05 [62.80, 64.44] & 67.59 [71.20, 71.32] \\
8/255 & 8/255 & 41.24 [54.01, 64.31] & 50.74 [56.63, 59.49] & 65.96 [67.63, 67.78] \\
8/255 & 16/255 & 36.96 [42.45, 52.90] & 39.15 [41.33, 44.22] & 58.17 [58.67, 58.85] \\
8/255 & 32/255 & 35.26 [37.13, 42.76] & 41.59 [43.01, 45.55] & 42.25 [43.02, 44.08] \\
\end{tabular}}
\caption{Error on clean data and (within brackets) under autoattack~\cite{autoattack_1, autoattack_2} and certified, for CIFAR10, $A^5/RC$, under attack $\epsilon_A = 8 / 255$. During training we use different $\epsilon_A^R$ and $\epsilon_D$.
The robustifier $R$ in the first row uses $\epsilon_D = 0$ and thus it does not provide any form of defense, whereas the corresponding robust CROWN-IBP classifier is fine tuned using $\epsilon_A^C = \epsilon_R$ during training.}
\label{tab:cifar10_a5rc}
\end{table}

\begin{figure*}
    \centering
    \begin{tabular}{l|c|c}
     Algo & Success & Failure \\
    \hline
    \rotatebox{90}{$A^5/O$} &
    \includegraphics[trim=6.5cm 14cm 4.5cm 2cm, clip, width=0.46\textwidth]{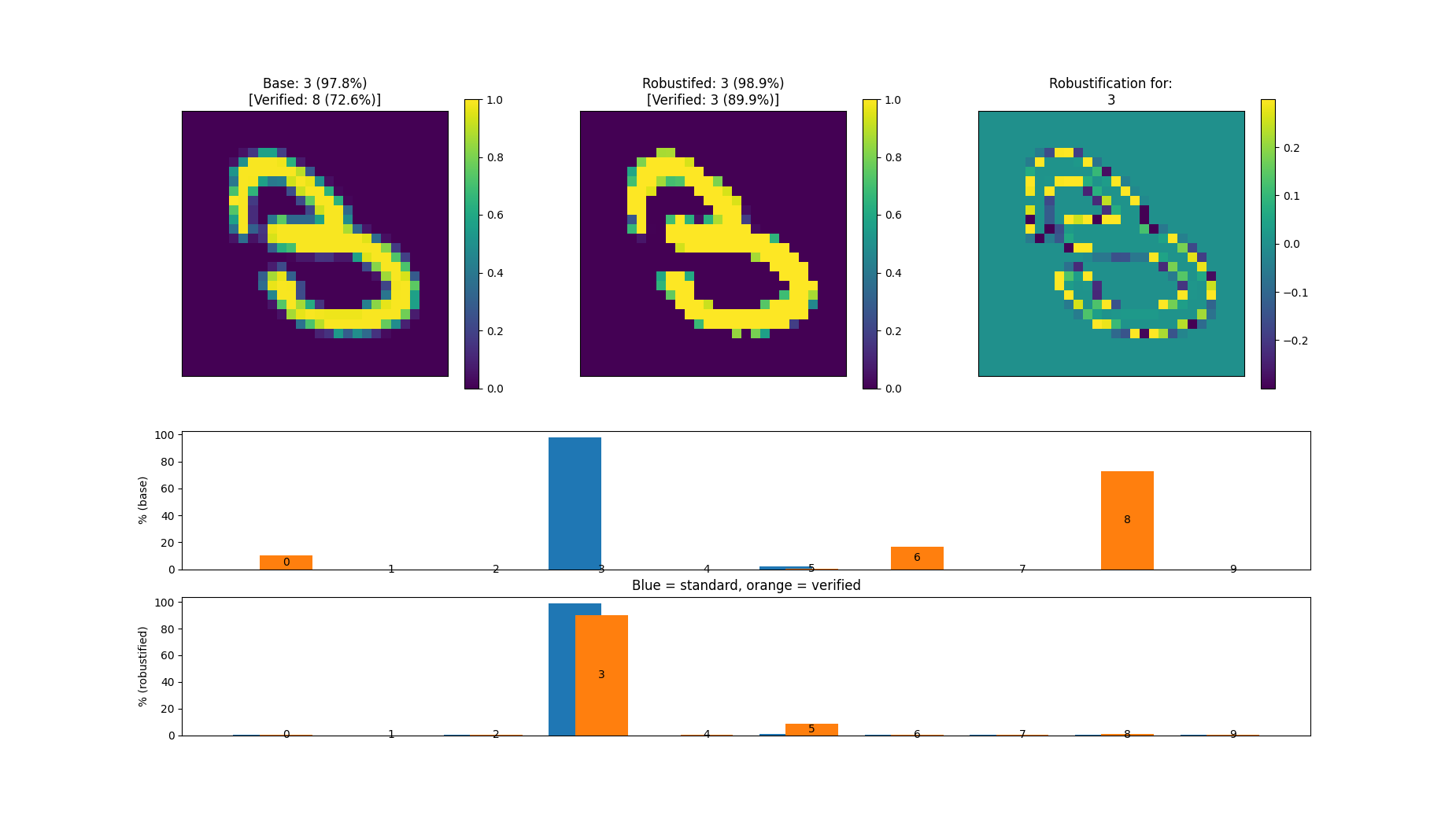} &
    \includegraphics[trim=6.5cm 14cm 4.5cm 2cm, clip, width=0.46\textwidth]{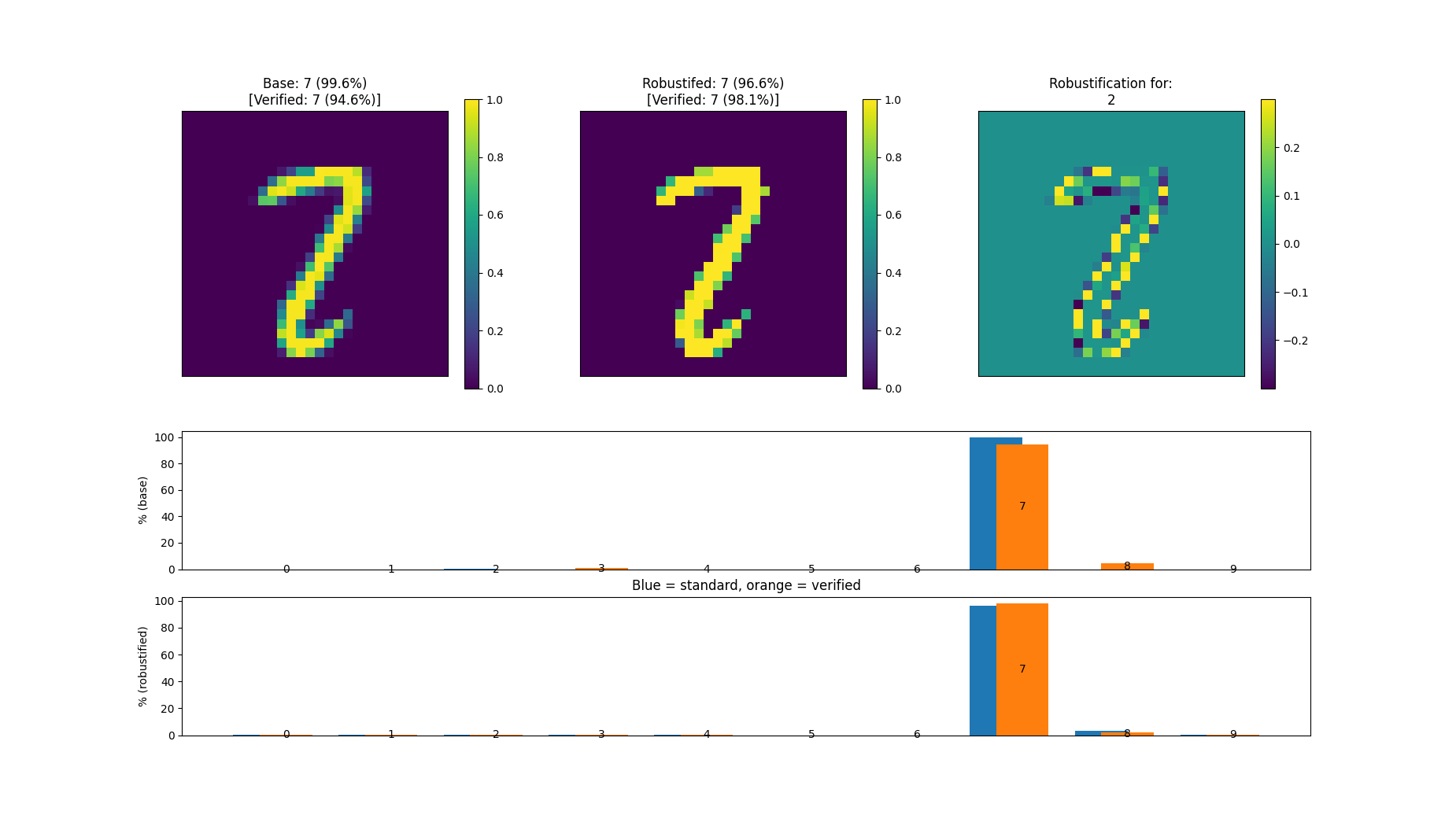} \\
    %\hline
    \rotatebox{90}{$A^5/R$} &
    \includegraphics[trim=6.5cm 14cm 14cm 2cm, clip, width=0.46\textwidth]{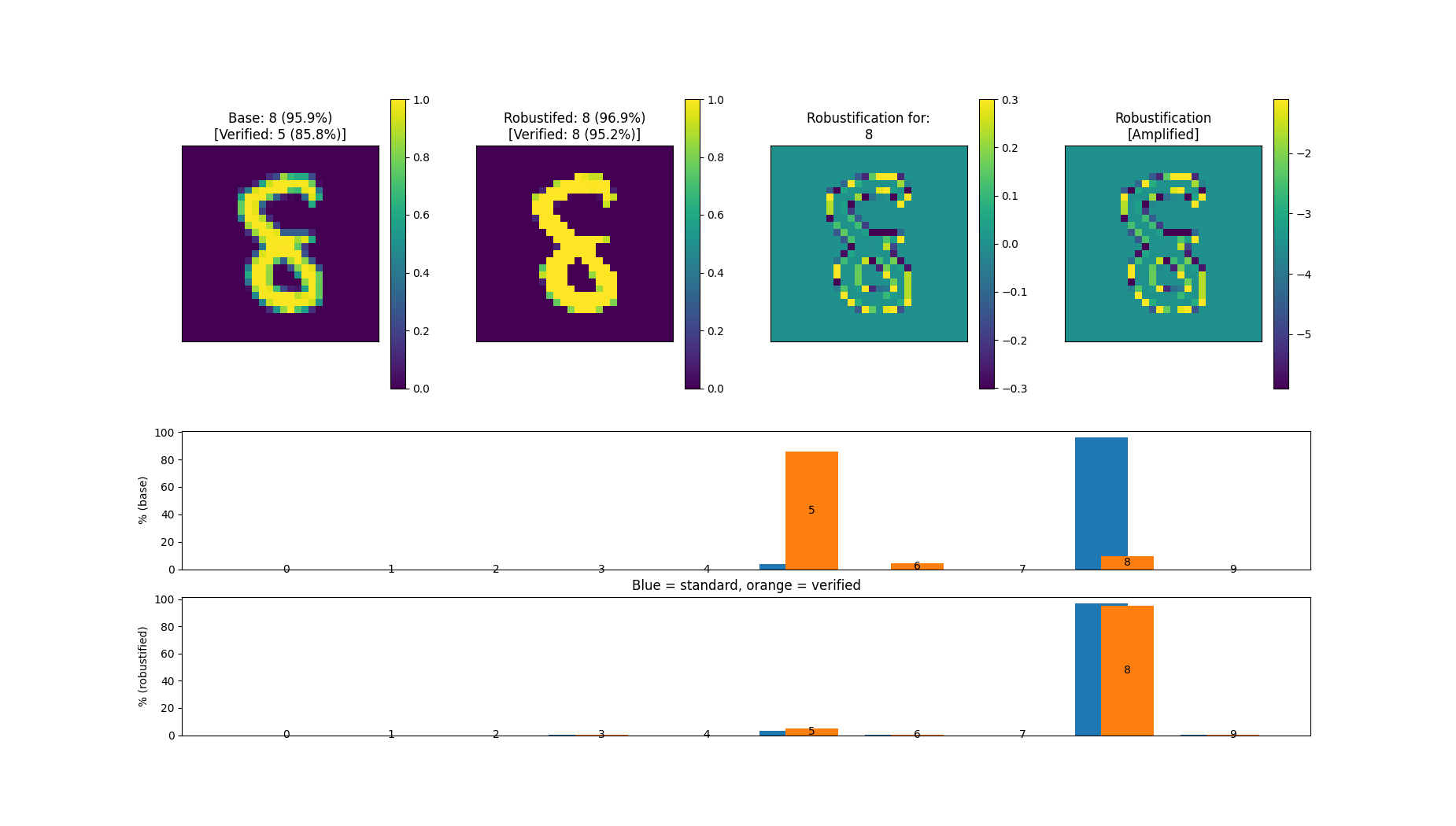} &
    \includegraphics[trim=6.5cm 14cm 14cm 2cm, clip, width=0.46\textwidth]{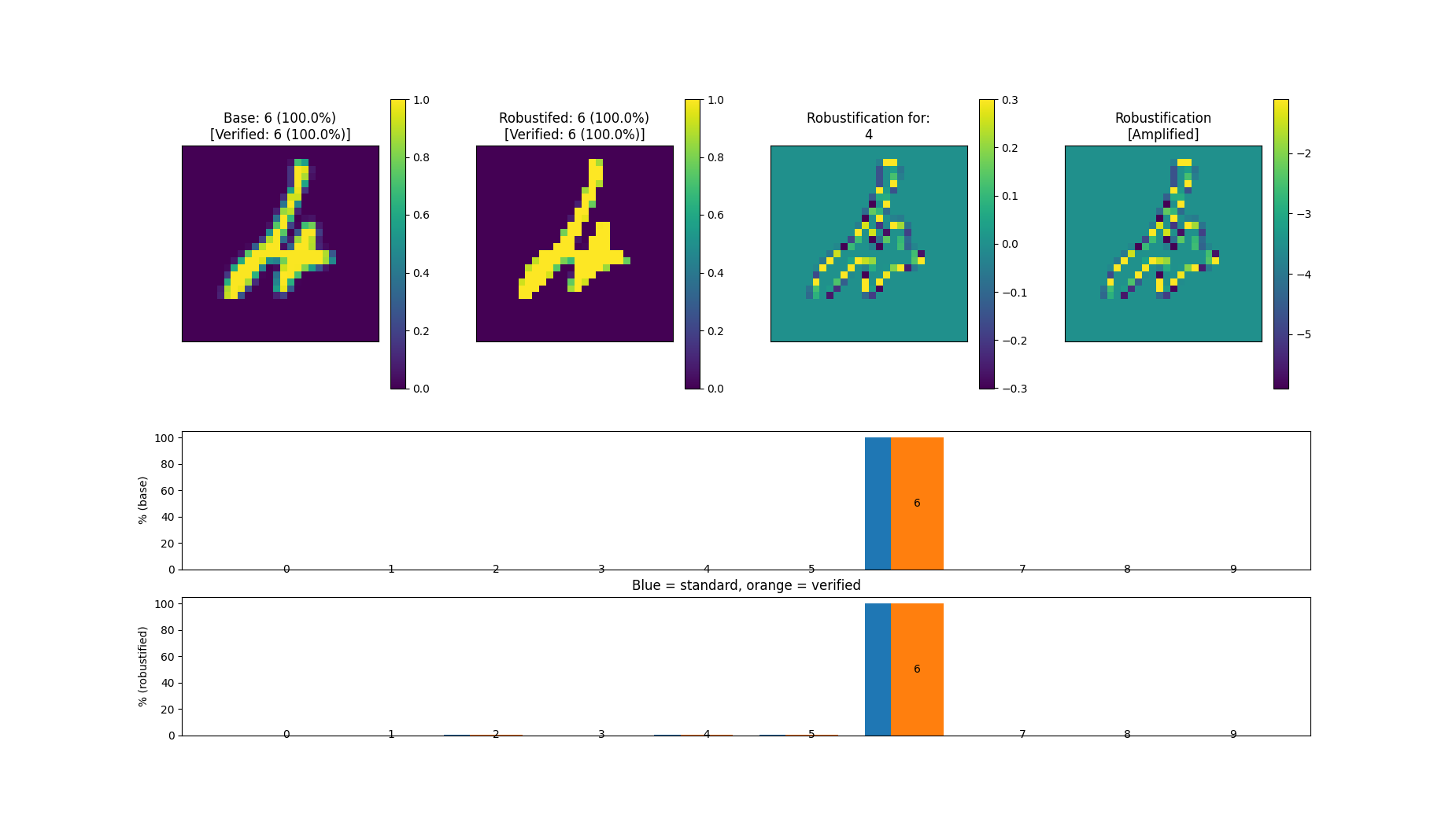} \\
    %\hline
    \rotatebox{90}{$A^5/RC$} &
    \includegraphics[trim=6.5cm 14cm 14cm 2cm, clip, width=0.46\textwidth]{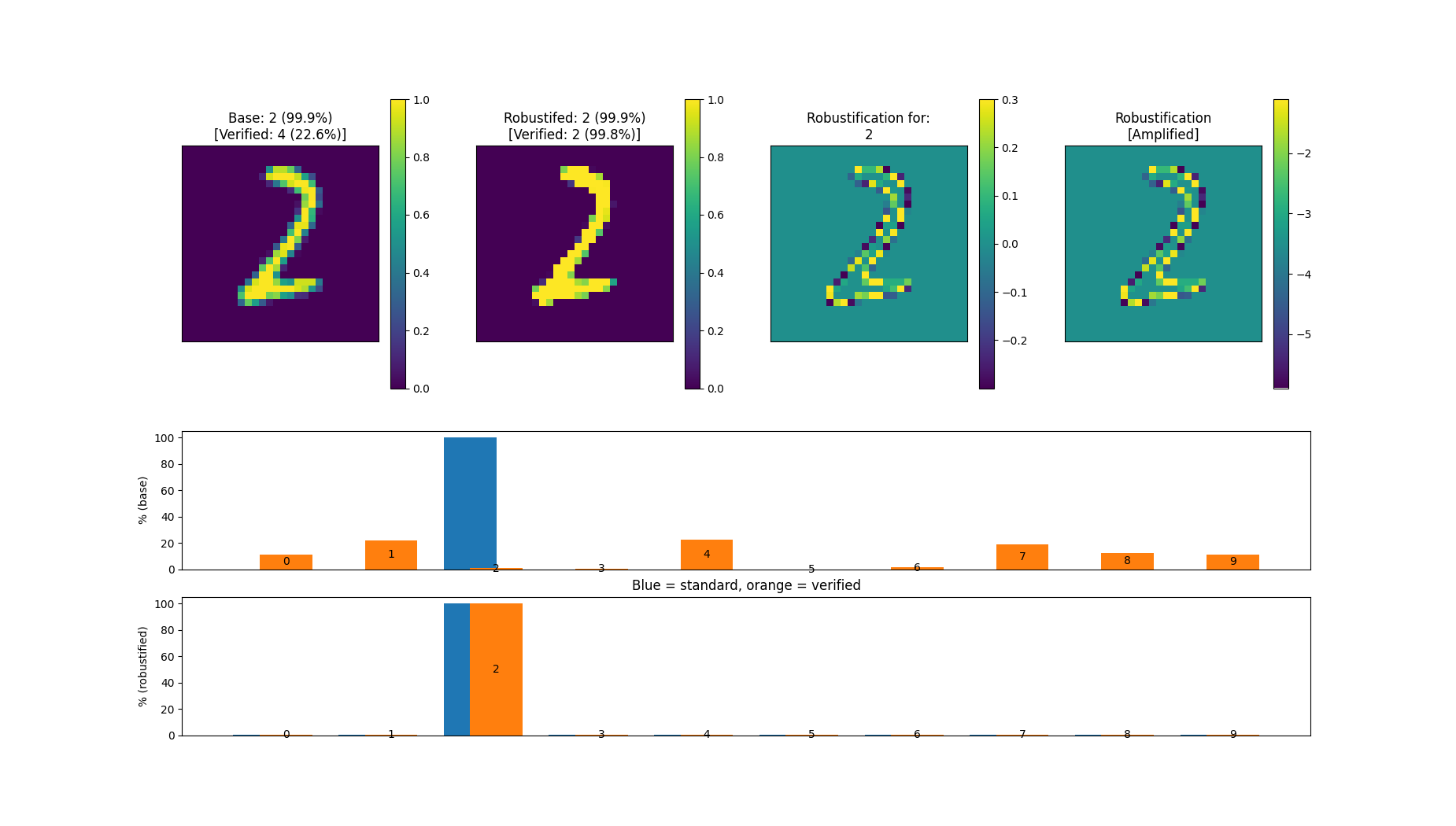}&
    \includegraphics[trim=6.5cm 14cm 14cm 2cm, clip, width=0.46\textwidth]{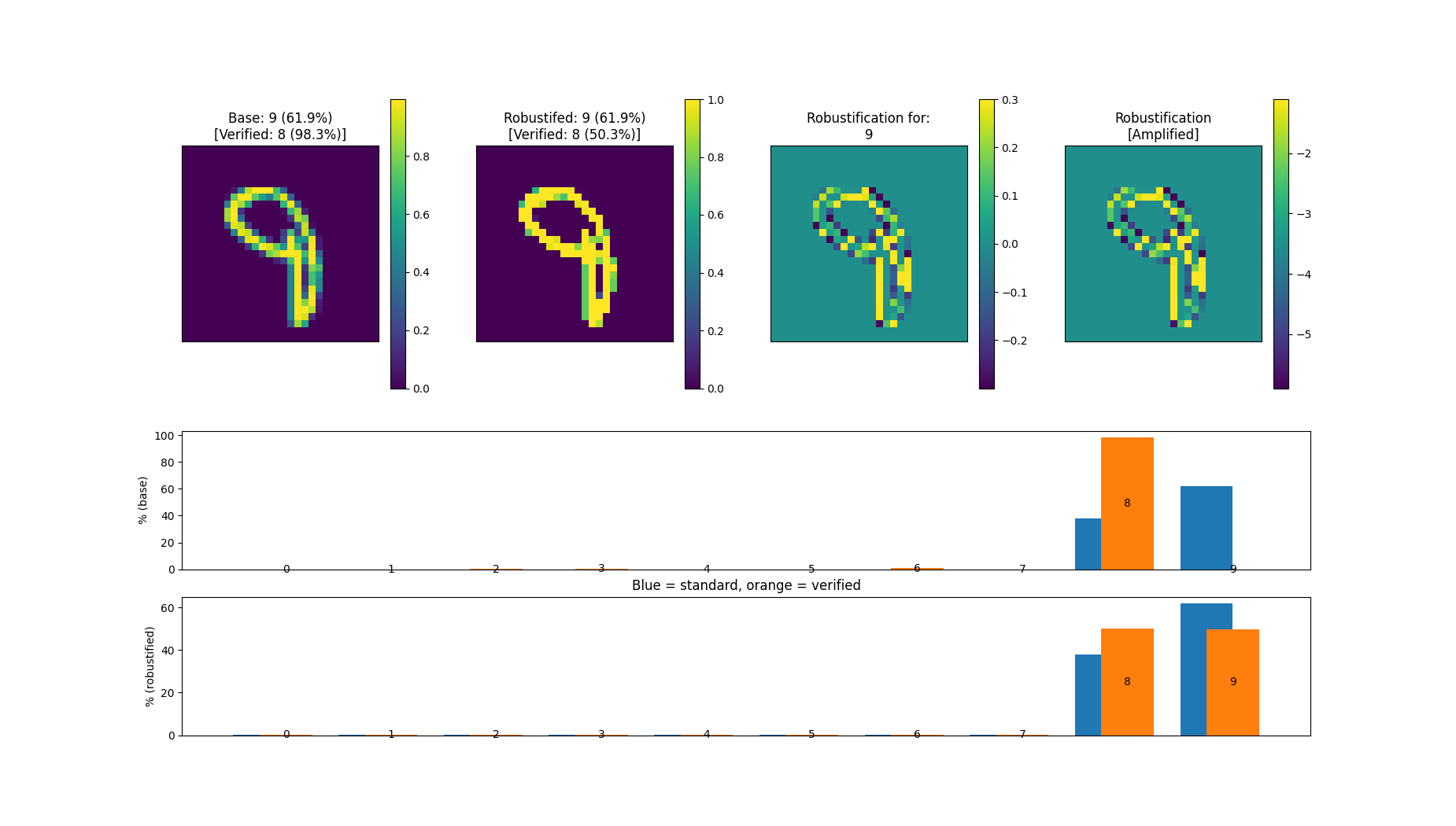} \\
    \hline
    %\hline
    \rotatebox{90}{~~~~$A^5/O$} &
    \includegraphics[trim=6.5cm 14cm 14cm 2cm, clip, width=0.46\textwidth]{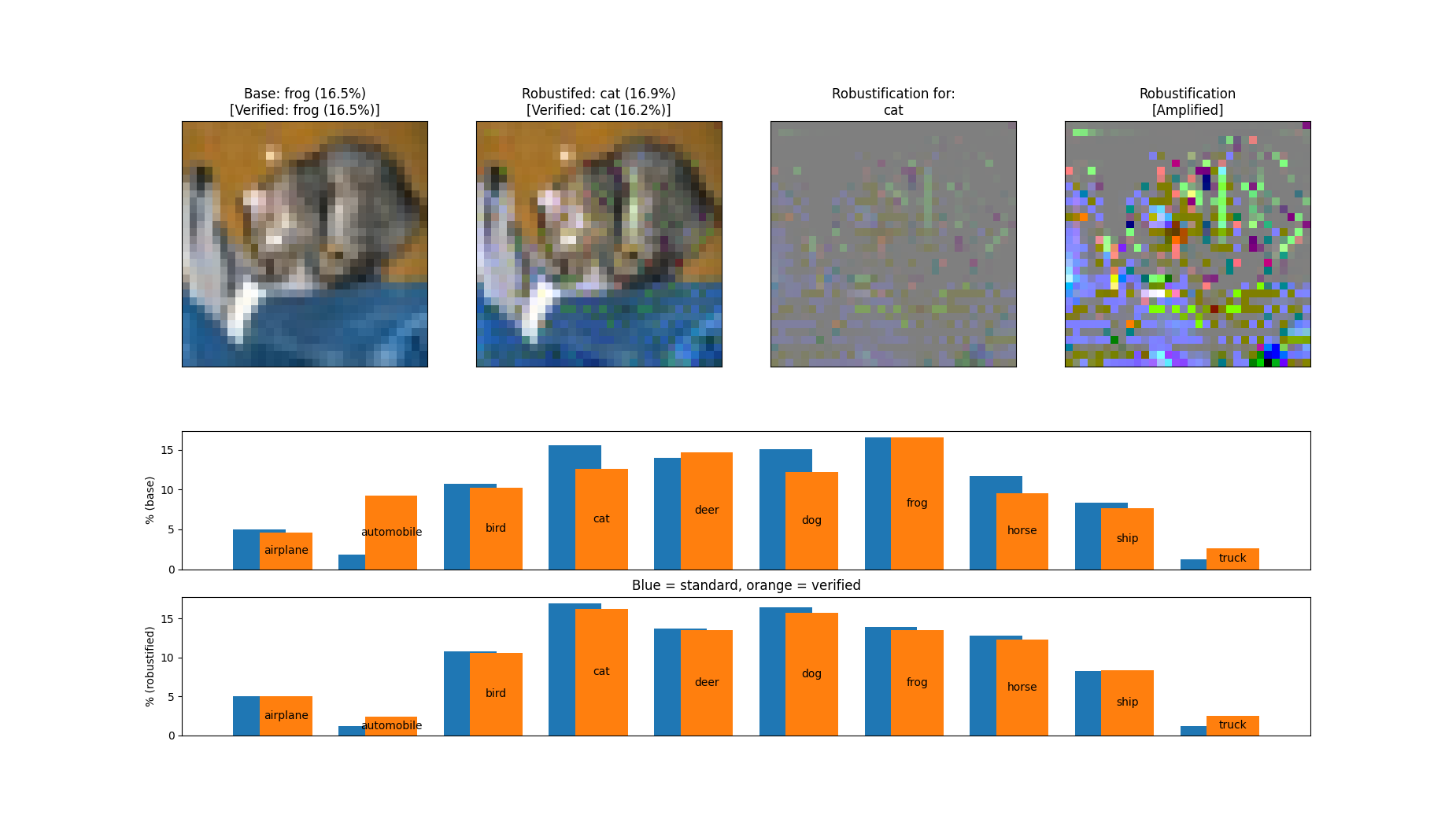} &
    \includegraphics[trim=6.5cm 14cm 14cm 2cm, clip, width=0.46\textwidth]{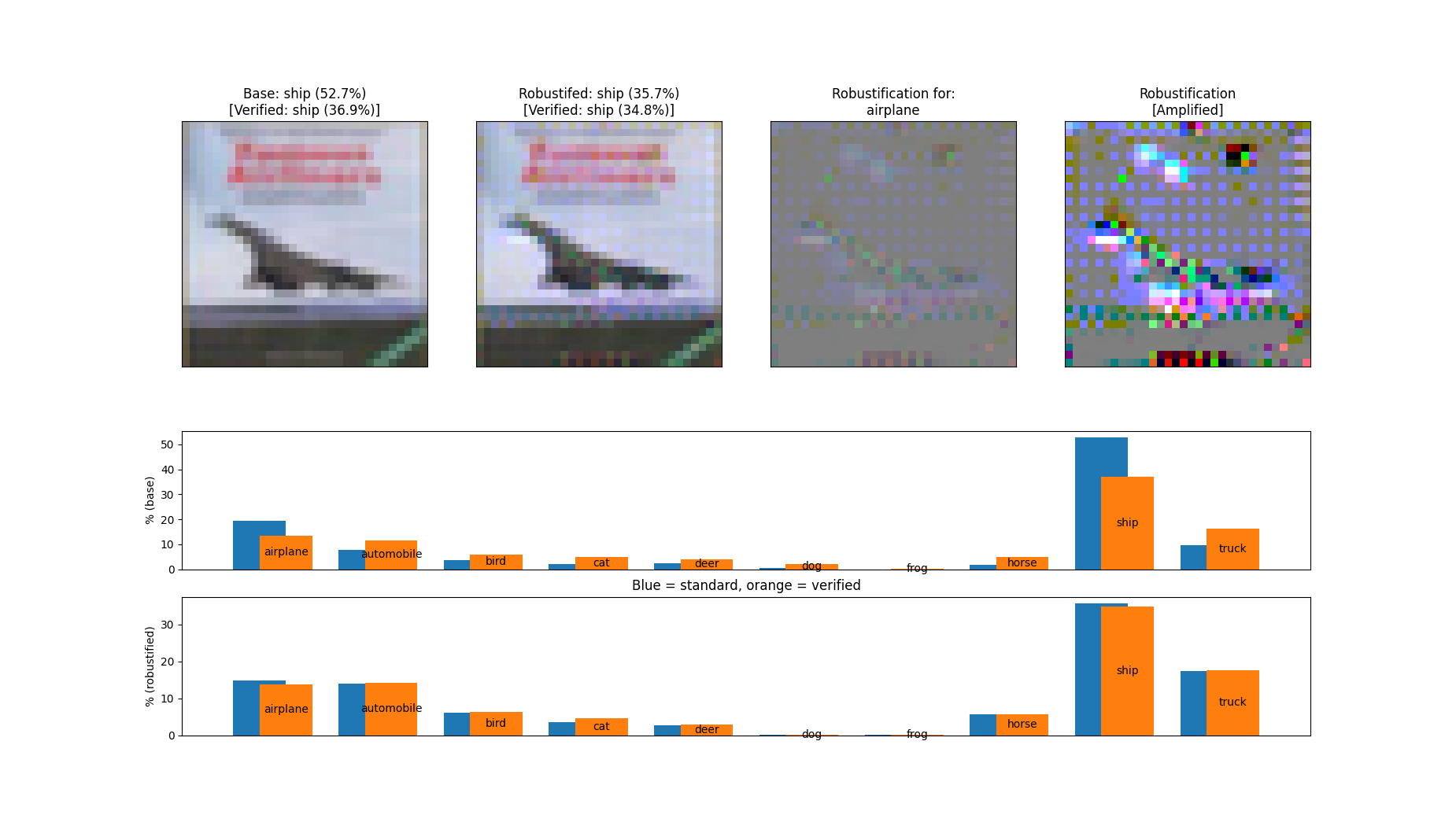} \\
    %\hline
    \rotatebox{90}{$A^5/R$} &
    \includegraphics[trim=6.5cm 14cm 14cm 2cm, clip, width=0.46\textwidth]{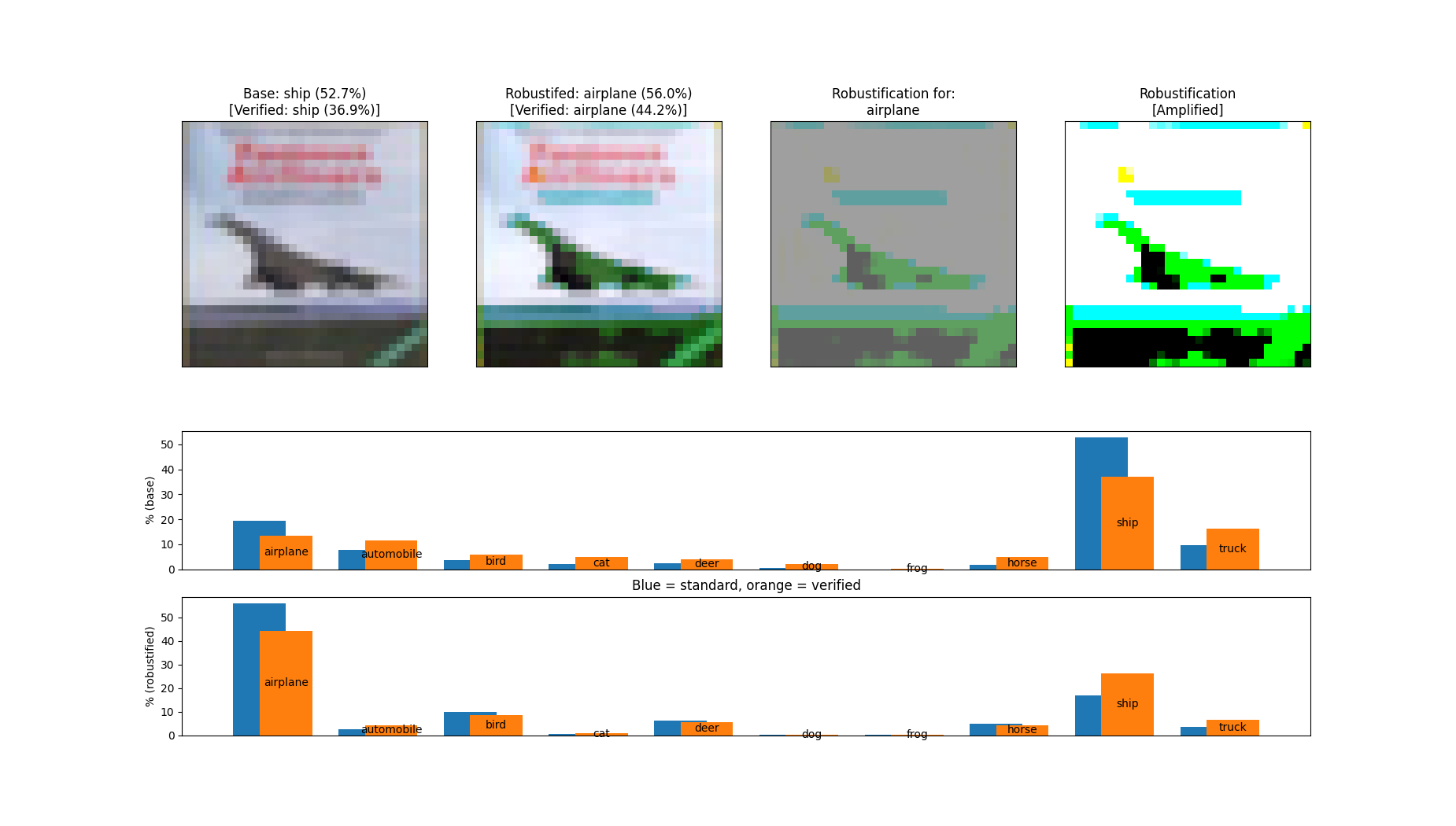} &
    \includegraphics[trim=6.5cm 14cm 14cm 2cm, clip, width=0.46\textwidth]{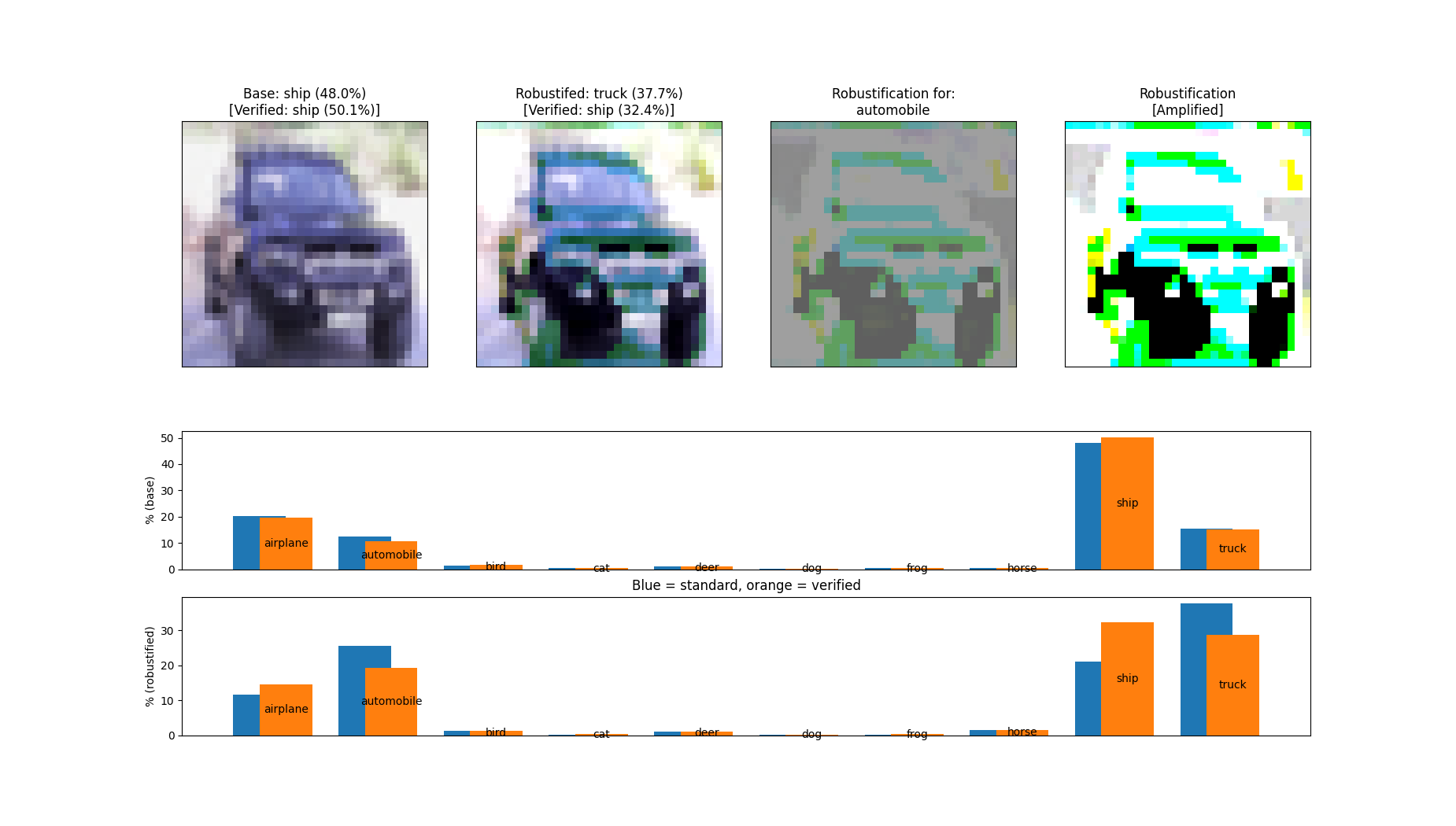} \\
    %\hline
    \rotatebox{90}{$A^5/RC$} &
    \includegraphics[trim=6.5cm 14cm 14cm 2cm, clip, width=0.46\textwidth]{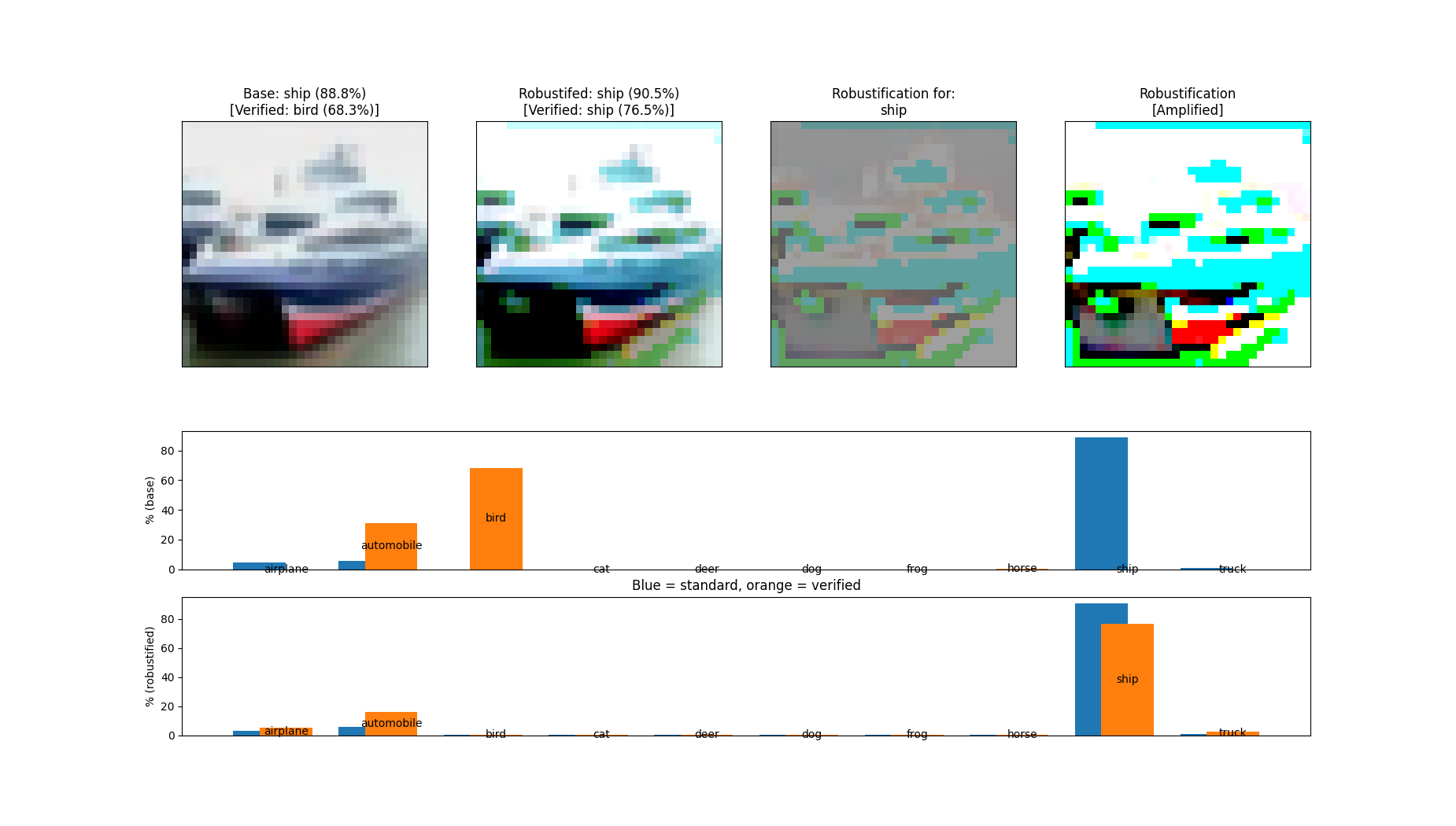}&
    \includegraphics[trim=6.5cm 14cm 14cm 2cm, clip, width=0.46\textwidth]{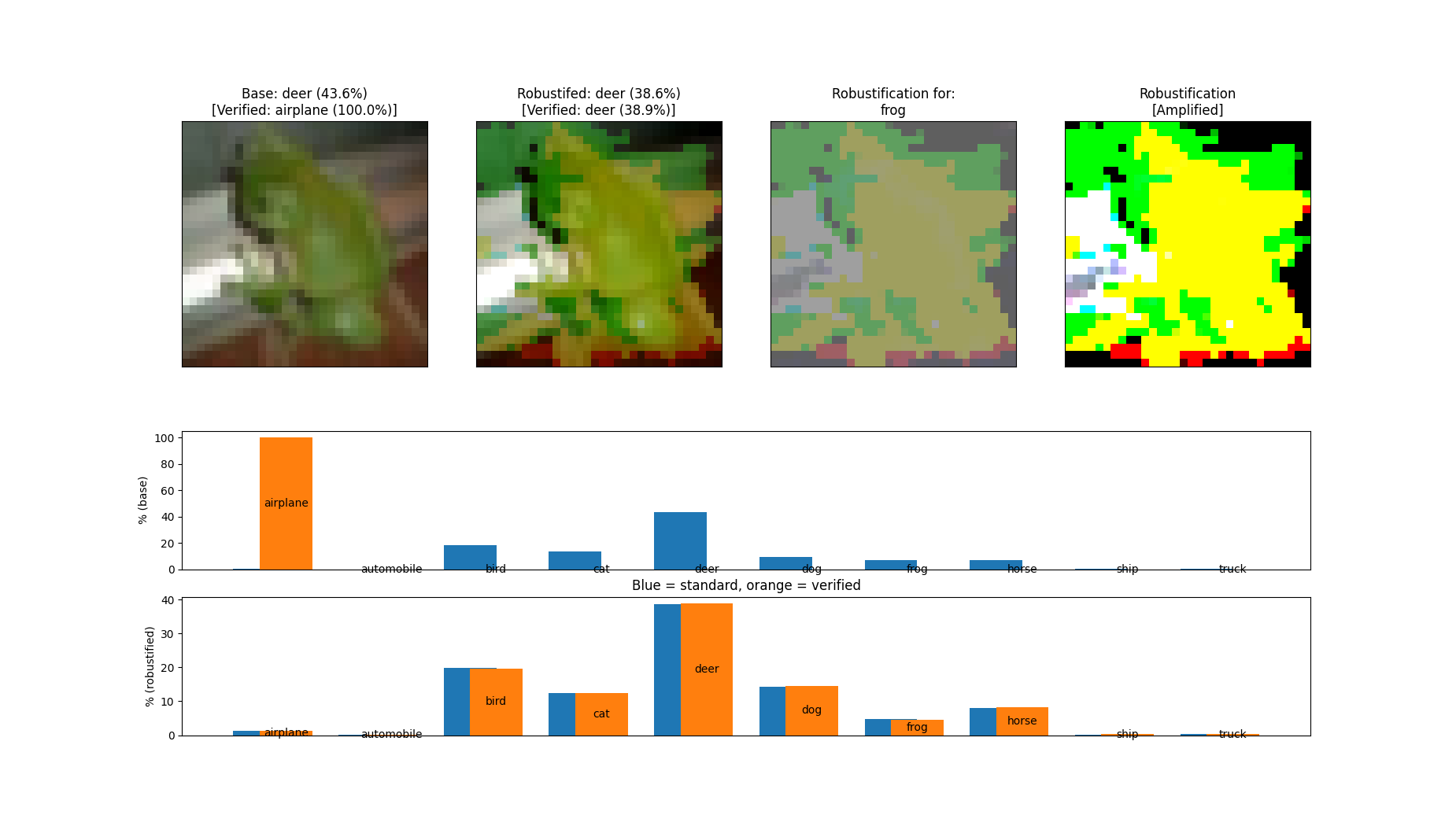} \\
    %\hline
    \end{tabular}
    \caption{Successful and failed robustifications
for $A^5/O$, $A^5/R$, and $A^5/RC$, for MNIST ($\epsilon_A = \epsilon_A^C = \epsilon_A^R = \epsilon_D = 0.3$) and CIFAR10 ($\epsilon_A=8.8/255$,  $\epsilon_A^C=8/255$, $\epsilon_A^R=4/255$, and $\epsilon_D=32/255$).
Each triplet shows the original (left) and robustified (center) images; the rightmost panel is the defensive augmentation.
All the $A^5$ recipes consistently increase the contrast and the high frequency content on MNIST, where failure cases are ambiguous even for a human observer. For CIFAR10, $A^5/O$ leverage isolated pixel changes, whereas $A^5/R$ and $A^5/RC$ consistently increase the color contrast and saturation, similarly to results in literature~\cite{preemptive_Moo22}.}
    \label{fig:examples}
\end{figure*}

Fig.~\ref{fig:examples} shows examples of robustification on MNIST.
At visual inspection, $A^5/O$, $A^5/R$, and $A^5/RC$ act similarly on these images: they add contrast and enhance the high frequency features (\ie, edges) of the digits.
The few failure cases are often pictures that look ambiguous even to a human observer (right panels in Fig.~\ref{fig:examples}).

\begin{figure*}
    \centering
    \includegraphics[width=\textwidth]{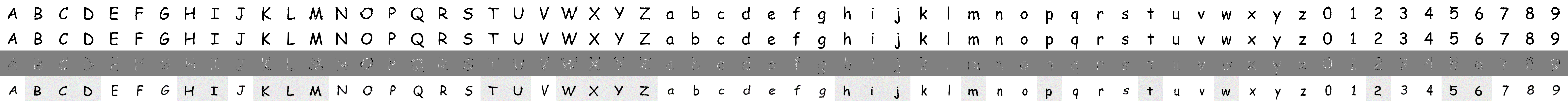}
\caption{The first, second and third row show respectively the robustified ($\boldsymbol{w}+\delta\boldsymbol{w}_D$), nominal ($\boldsymbol{w}$) and robustification ($\delta\boldsymbol{w}_D$) sets of 62 letters considered for testing $A^5/PC$. The last row show examples of the robustified letters after framing them, $\boldsymbol{x}+\delta\boldsymbol{x}_D = A(\boldsymbol{w}+\delta\boldsymbol{w}_D)$.}
    \label{fig:ocr}
\end{figure*}

We perform tests on CIFAR10 and compare $A^5$ against a state of the art CROWN-IBP classifiers $C$, trained with $\epsilon_A^C=8.8/255$, that we also adopt as legacy $C$ for $A^5$.
To test the effect of the magnitude of the attack used for robustification (with $A^5/O$) or for training $R$ (for $A^5/R$, $A^5/RC$), we use $\epsilon_A^R=\{4/255, 8/255, 16/255\}$ that are smaller, equal or greater than $\epsilon_A^C$.
Results in Table~\ref{tab:cifar10_a5o} are consistent with those measured for MNIST: $A^5/O$ significantly improves the clean and certified error rates over the base CROWN-IBP classifier.
Training $R$ through $A^5/R$ produces a smaller improvement, still sufficient to beat the CROWN-IBP classifier by a significant margin (Table ~\ref{tab:cifar10_a5r}).
The value of $\epsilon_A^R$ has a significant impact on the final result: if too large ($\epsilon_A^R > \epsilon_A$), the clean accuracy is penalized, because $A^5$ looks for very large plateaus in the classification landscape, that are hard or impossible to find.
In other words, we observe for $A^5$ a trade-off between the clean and certified accuracy, likewise adversarial training and certified methods.
Best results on CIFAR10 are achieved again by $A^5/RC$, when $\epsilon_A^R$ is equal or slightly smaller than $\epsilon_A = \epsilon_A^C$.
The improvements in clean and certified error are in this case in the impressive range of $19\%$ and $24\%$ respectively (Table~\ref{tab:cifar10_a5rc}).
We also notice here that the theoretical certified error rates measured with auto{\_}LiRPA~\cite{autolirpa_Xu21} in $A^5$ and those measured experimentally by autoattack~\cite{autoattack_1, autoattack_2} are close for $A^5$, whose working point at convergence may favor the consistency between these two estimates.

Also for CIFAR10 we perform a visual inspection of defensive augmentation (Fig.~\ref{fig:examples}).
$A^5/O$ exploits a slight contrast increase and modifies isolated pixels: likewise adversarial attacks with small $p$ norm (up to single pixel attacks~\cite{OnePixel_Su19}), the same seems possible when crafting defensive perturbations.
Preemptive robustification~\cite{preemptive_Moo22} shows a similar pattern (see their Fig. 6): non natural textures and enhanced color saturation emerge in their images for large defensive perturbations.
The robust images crafted by $R$ in $A^5/R$ or $A^5/RC$, however, are different: they generally show  increased contrast and color saturation, without the textures generated by $A^5/O$ and~\cite{preemptive_Moo22}. 
We speculate that these textures may be associated with the iterative optimization in $A^5/O$ and~\cite{preemptive_Moo22}, that may both suffer some form of obfuscated gradient~\cite{obfuscation_Ath18}; a more regular defense is instead generated by $R$ without resorting to any iterative process.

\begin{table}
%\begin{wraptable}{r}{7cm}
    \centering
    \begin{tabular}{c|c}
         Algo &  Error [certified error] \\
         \hline
         Vanilla & 0.89\% [100.00\%] \\
         CROWN-IBP & 3.85\% [13.85\%] \\
         \hline
         $A^5/P$ & 3.08\% [11.84\%] \\
         $A^5/PC$ & 0.73\% [4.20\%]
    \end{tabular}
    \caption{Error and certified error for the classification of the 62 characters in Fig.~\ref{fig:ocr}, for $\epsilon_A = 0.1$ and different algorithms. $A^5/P$ changes the physical shape of the letters to achieve robustness after framing them; $A^5/PC$ also trains the associated classifier.}
    \label{tab:ocr}
%\end{wraptable}
\end{table}

Additional results on FashionMNIST and Tinyimagenet, overall consistent with the ones presented here and showing the quantitative advantages provided by $A^5$ on these datasets, are reported in the Appendix.
These experiments highlight two important facts.
The first one is that the integration of the design philosophy of $A^5$ with that of other recent ideas in the field of adversarial defense (like the $\ell_\infty$-dist neurons that implement 1-Lipschitz functions and are inherently robust to adversarial attacks) may lead to even bigger improvements in terms of certified robustness.
The second is that the integration of different defense strategies may indeed be strictly needed to guarantee further progresses, as scaling to very large dataset remains problematic even for a preemptive robustification algorithms like $A^5$.

Finally, for $A^5/P$ and $A^5/PC$, we test the classification of an alphabet of 62 characters (Fig.~\ref{fig:ocr}), after random rotation, shift, perspective distortion, noise addition, blurring and color jittering (details in Appendix), that simulate the scanning of a document for OCR.
Table~\ref{tab:ocr} shows the clean and certified errors for: a vanilla, non robust classifier; a CROWN-IBP classifier; $A^5/P$; and $A^5/PC$.
$A^5$ achieves again a significant improvement both on clean and certified errors: $A^5/PC$ has better ($0.73\%$) accuracy of the vanilla $C$ ($0.89\%$) and can be attacked in only $4.20\%$ of the cases.
This results is obtained with a slight modification of the shape of the prototype characters shown in Fig.~\ref{fig:ocr}.

\section{Discussion and conclusion}
\label{sec:conclusion}

We introduce $A^5$, the first framework that leverages adversarial augmentation to preemptively modify the input of a DNN or a physical object, to make it certifiable robust against adversarial attacks.
$A^5$ is not simply complementary to other defense methods: its tights with them are strict and they can be used together to achieve higher robustness. For instance, one of the limitations we found is that $A^5/R$ requires a somehow robust initial classifier $C$.
Co-training $R$ and a poorly robust $C$ easily compensates for the initially low robustness of $C$ and boosts its high clean accuracy while also achieving state of the art results in terms of certified accuracy; finding the optimal initial robustness of $C$ remains an open question though.

Here we do not investigate all the possible $A^5$ recipes: in its most general implementation ($A^5/PRC$) the physical objects, $R$ and $C$ could all be optimized at the same time.
This opportunity can be explored through the released code at \url{https://github.com/NVlabs/A5}.
This will also serve answering other questions that do not find space here: for instance, we do not know if robustification generalizes to other classifiers as adversarial attacks do, how $A^5$ works in norms other than $p=\infty$, and how much it can further improve if coupled with state-of-the-art solutions like the recently proposed $\ell_{\infty}$-dist neurons~\cite{Zha21_LInf}.

Different $A^5$ recipes find applications in scenarios where the user can control the acquisition device, equip it with $R$ and guarantee its protection (\eg{}, image acquisition on a phone before server communication) or while designing the infrastructure (\eg{}, robust road signs creation); it may be not suitable for more general adversarial scenarios.
Overall, we believe that the practical deployment of robust systems may benefit from (or even require) methods for both robust classifiers and preemptive robustification.

%Finally, although we tested on CIFAR10, we still need to test $A^5$ on full resolution, real images.

%As for the potential societal impacts of our work, we believe that $A^5$ may be beneficial for the development of secure artificial intelligent systems --- at the same time, publishing defense techniques also provides information to the attackers. In the longer term, research in this field may lead to the development of novel infrastructures (or change to the existing ones) that are inherently robust (\eg, road signs); this may be seen both as an opportunity and as a cost for the society at the same time.

% For patenting only: A last note to be made here is the connection between$A^5$ and style transfer: when creating robustified fonts, they are not very regular - is there a better, more controlled way to perform robustification, which is then similar to style transfer (e.g. add inclination to the font).

%PUT ME SOMEWHERE It is also worthy noticing that, with the aim of preserving the natural appearance of the robust objects, one could work with constraints different from the norm constraint ($||\delta\boldsymbol{x}_D||_\infty < \epsilon_D$) adopted here; for instance, in the OCR case, one could control the rotation or thickness of the each character to increase their robustness, while also keeping their original appearance.
\section{Appendix}
\label{sec:supplementary}

\subsection{Threat models and usage scenarios}

\begin{figure*}[ht!]
\centering
\setlength{\tabcolsep}{1pt}
\begin{tabular}{c|c|c}
Recipe & Training time & Run time \\
\hline
\begin{tabular}{c}$A^5$\end{tabular} & 
\begin{tabular}{c}
\includegraphics[width=0.55\textwidth,clip=True, trim=0cm 0cm 2.2cm 0cm]{figs/recipes/Slide1.pdf} \end{tabular} &  \begin{tabular}{c} ~ \end{tabular}\\
\multicolumn{3}{l}{\footnotesize{\textbf{Threat  model / Training notation / Usage and notes.} General framework, it encompasses all the following recipes.}} \\
\midrule
\begin{tabular}{c}$A^5/O$\end{tabular} & \begin{tabular}{c}\includegraphics[width=0.55\textwidth,clip=True, trim=0cm 2.5cm 2.5cm 0cm]{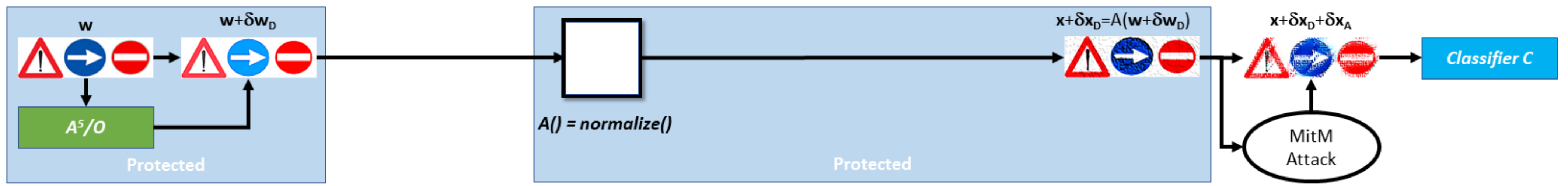} \end{tabular} & \begin{tabular}{c}\includegraphics[width=0.30\textwidth,clip=True, trim=0cm 2.5cm 23cm 0cm]{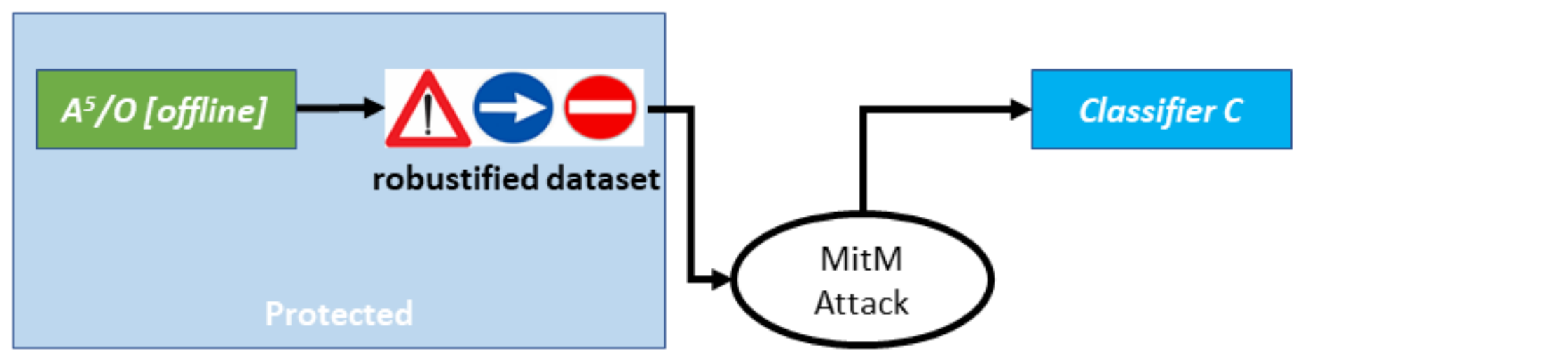} \end{tabular}\\
\multicolumn{3}{l}{\footnotesize{\textbf{Threat  model.} Robustify a dataset (offline) against white/black box, MitM attacks.}} \\
\multicolumn{3}{l}{\footnotesize{\textbf{Training notation.} $\boldsymbol{w}$: dataset. $\boldsymbol{x}$: normalized dataset.}} \\
\multicolumn{3}{l}{\footnotesize{\textbf{Usage and notes.} No practical use, apart from studying $A^5$. Use the legacy classifier $C$.}}\\
\midrule
\begin{tabular}{c}$A^5/R$\end{tabular} & \begin{tabular}{c}\includegraphics[width=0.55\textwidth,clip=True, trim=0cm 2.5cm 2.5cm 0cm]{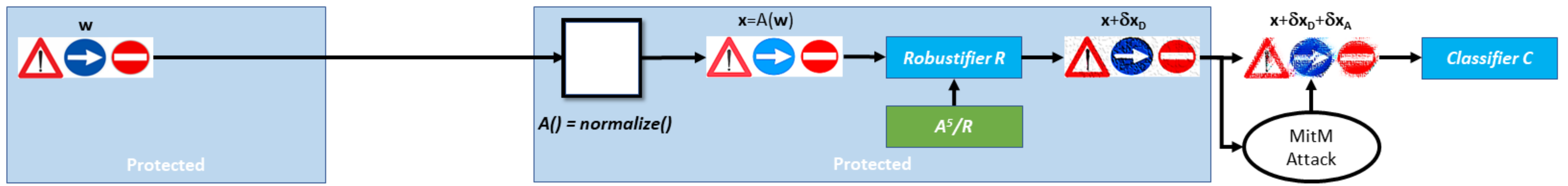} 
\end{tabular} & \begin{tabular}{c}\includegraphics[width=0.30\textwidth,clip=True, trim=0cm 2.5cm 23cm 0cm]{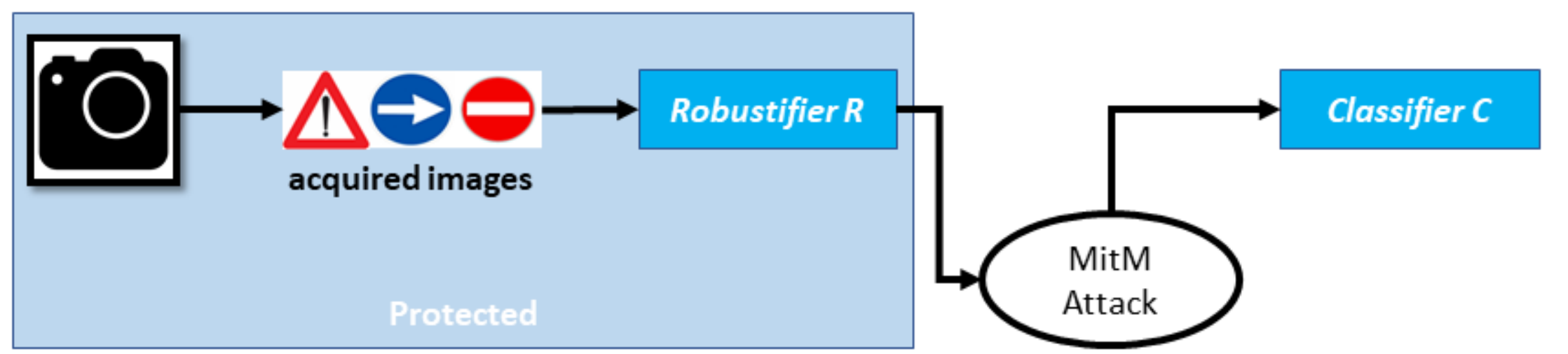} \end{tabular} \\
\begin{tabular}{c}$A^5/RC$\end{tabular} & \begin{tabular}{c}\includegraphics[width=0.55\textwidth,clip=True, trim=0cm 2.5cm 2.5cm 0cm]{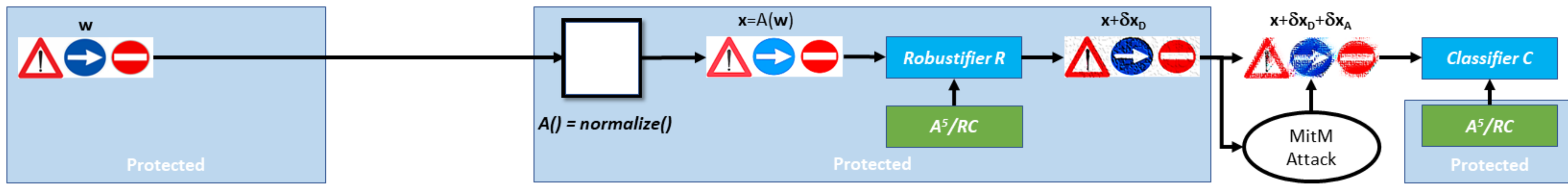} \end{tabular} & \begin{tabular}{c}\includegraphics[width=0.30\textwidth,clip=True, trim=0cm 2.5cm 23cm 0cm]{figs/recipes/Slide8.pdf} \end{tabular} \\
\multicolumn{3}{l}{\footnotesize{\textbf{Threat  model.} Robustify acquired data on-the-fly (using a robustifier $R$) against white/black box, MitM attacks.}} \\
\multicolumn{3}{l}{\footnotesize{\textbf{Training notation.} $\boldsymbol{w}$: dataset. $\boldsymbol{x}$: normalized dataset.}} \\
\multicolumn{3}{l}{\footnotesize{\textbf{Usage and notes.} Acquired data $\boldsymbol{x}$ are passed to the robustifier $R$ on-the-fly. $A^5/R$ uses the legacy classifier $C$, $A^5/RC$ retrains it.}}\\
\midrule
\begin{tabular}{c}$A^5/P$\end{tabular} & \begin{tabular}{c}\includegraphics[width=0.55\textwidth,clip=True, trim=0cm 2.5cm 2.5cm 0cm]{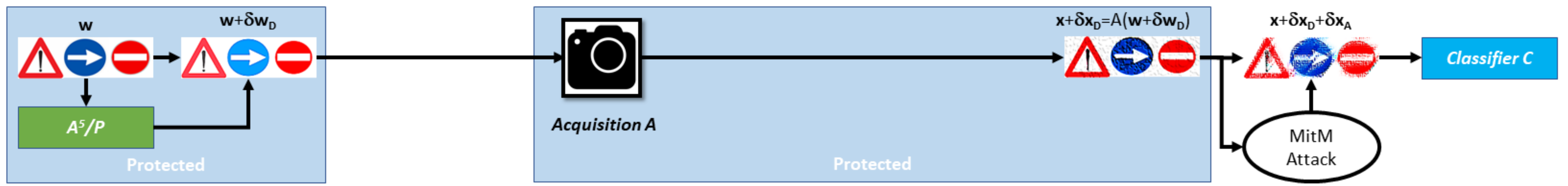} \end{tabular} & \begin{tabular}{c}\includegraphics[width=0.30\textwidth,clip=True, trim=0cm 2.5cm 23cm 0cm]{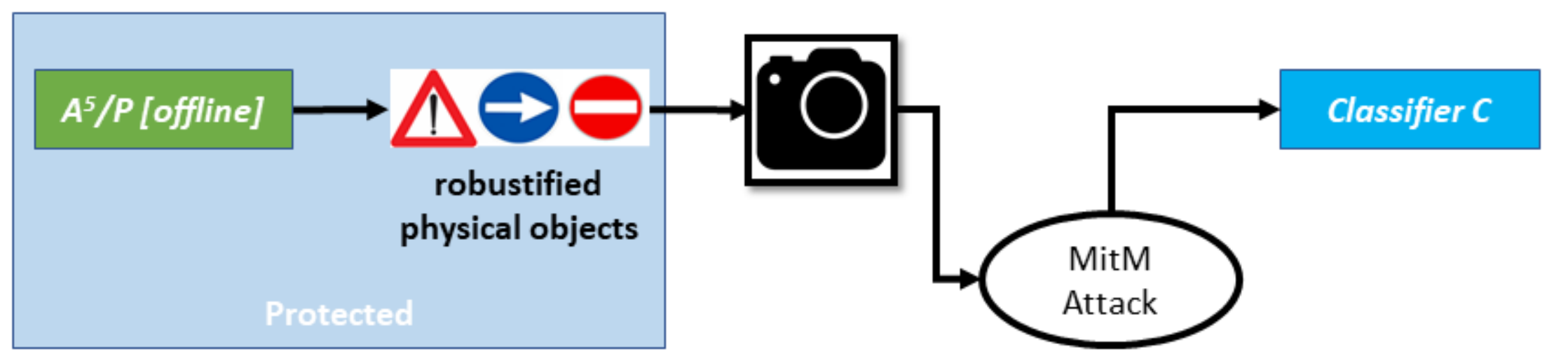} \end{tabular} \\
\begin{tabular}{c}$A^5/PC$\end{tabular} & \begin{tabular}{c}\includegraphics[width=0.55\textwidth,clip=True, trim=0cm 2.5cm 2.5cm 0cm]{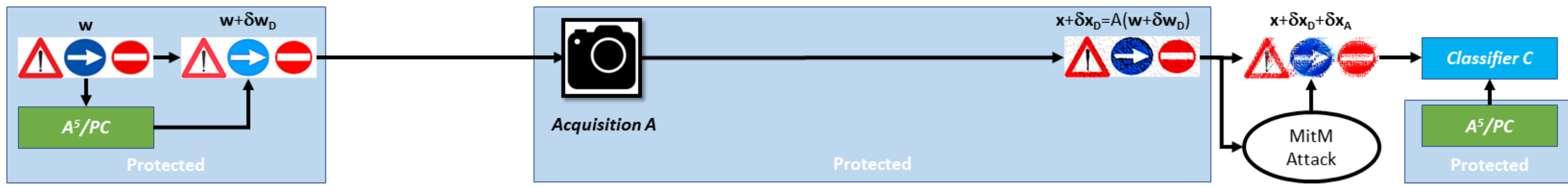} \end{tabular} & \begin{tabular}{c}\includegraphics[width=0.30\textwidth,clip=True, trim=0cm 2.5cm 23cm 0cm]{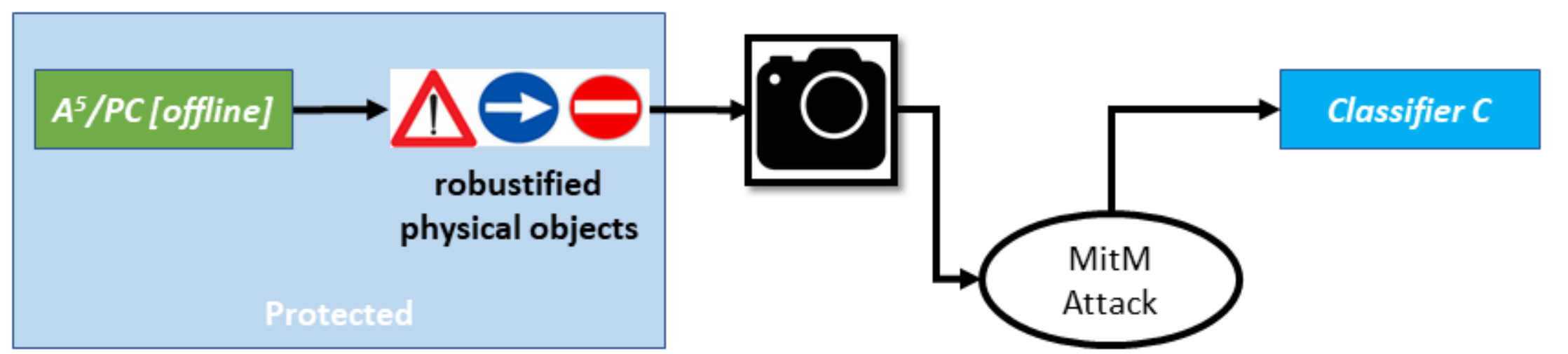} \end{tabular} \\
\multicolumn{3}{l}{\footnotesize{\textbf{Threat  model.} Robustify physical objects (offline) against white/black box, MitM attacks against the acquired images of the same objects.}} \\
\multicolumn{3}{l}{\footnotesize{\textbf{Training notation.} $\boldsymbol{w}$: physical objects. $\boldsymbol{x}$: normalized acquired images.}} \\
\multicolumn{3}{l}{\footnotesize{\textbf{Usage and notes.} Acquired data $\boldsymbol{x}$ are passed to the robustifier $R$ on-the-fly. $A^5/R$ uses the legacy classifier $C$, $A^5/RC$ retrains it.}}\\
\midrule
\end{tabular}
\caption{The first row is a schematic representation of the full $A^5$ framework. It allows the robust augmentation of both physical objects $\boldsymbol{w}$ and acquired data $\boldsymbol{x}$ to make them certifiably robust against MitM and physical (not tested here) adversarial attacks. The remaining rows illustrate the $A^5$ recipes tested in our paper. The ground truth class is always known at training time, whereas it is unknown at run time.}
\label{fig:A5_complete}
%\end{wrapfigure}
\end{figure*}

The first row in Fig.~\ref{fig:A5_complete} depicts a complete schematic representation of $A^5$, our framework for robust augmentation of both physical objects $\boldsymbol{w}$ or data $\boldsymbol{x}$, that makes them certifiably robust against MitM and physical (not tested here) adversarial attacks.
%Physical objects can be attacked by physical attacks $\delta\boldsymbol{w}_A$ (\eg, adversarial patches~\cite{AdvPAtch_Aba17}).
The $A^5$ model includes an acquisition block, that can simulate image capturing with a camera or perform simple data normalization depending on the threat and training scenario.
The acquired images $\boldsymbol{x}$ can be made certifiably robust against MitM attacks $\delta\boldsymbol{x}_A$ (crafted during their transmission to the classifier $C$) by a robustifier DNN $R$ that runs on-the-fly within a protected environment inaccessible to the attacker ($A^5/R$ and $A^5/RC$).
We show also the use of $A^5$ for offline robustification of datasets ($A^5/O$) and of physical objects $\boldsymbol{w}$ ($A^5/P$ and $A^5/PC$).
The green blocks in Fig.~\ref{fig:A5_complete} indicate the various $A^5$ recipes that are used to train $R$ and $C$ or to create robust physical objects or datasets.

$A^5$ can be used in a \emph{white-box scenario}  where the attacker has full access to the classifier $C$.
However, since $A^5$ uses certified bounds, it is agnostic to the specific nature of the attack: we can certify protection against any MitM attack $\delta \boldsymbol{x}_A$, $||\delta \boldsymbol{x}_A|| < \epsilon_A$ in Fig.~\ref{fig:A5_complete}. 
In practice, the attacker can use any white-, grey- or black-box algorithm to generate the adversarial attack.

Notice that MitM attacks can be crafted while transmitting the dataset to the classifier $C$. In a first scenario ($A^5/O$ in Fig.~\ref{fig:A5_complete}), we assume that $A^5$ runs offline, in a protected environment that is not accessible to the attacker, to preemptively robustify the samples of the dataset.
This situation is however mostly of theoretical and little practical interest.

In a second scenario ($A^5/R$ and $A^5/RC$ in Fig.~\ref{fig:A5_complete}), we assume a robustifier $R$ running on the acquisition device, in a protected environment; in this case robustification is performed on-the-fly,  before transmitting the data to $C$ (when data can be corrupted by a MitM attack) and without knowing the ground truth class.
This case can find practical application in several fields including automotive~\cite{Wan21_maninthemiddle} or anytime the capturing device communicates with another machine~\cite{Yan22_RobustSense}, under the assumption that $R$ runs together with the acquisition device in a protected environment: the attacker may have full knowledge of $R$ and $C$, but cannot modify $\boldsymbol{x}$ (before its transmission) or any internal state of the robustifier $R$. 

In a third scenario, $A^5$ furnishes protection against MitM attacks by creating certifiably robust physical objects ($A^5/P$ and $A^5/PC$ in Fig.~\ref{fig:A5_complete}).
In this case, the attacker may observe (but not interfere with) the object robustification process, that
is performed offline, in a protected environment.
This is in practice a very mild constraint, as robustification coincides with the physical creation of the objects. 
Practical examples are the design of certifiably robust road signs for automotive, fonts for OCR, or even audio signals~\cite{Qin19_audiomitm, Yak19_audiophys, Son21_preprocessingaudio}.

Notice that, in its more general formulation (first row in Fig.~\ref{fig:A5_complete}), $A^5$ can also protects against physical adversarial attacks like adversarial patches~\cite{AdvPAtch_Aba17} stick on top of a road sign, or adversarial sounds emitted in the environment to attack speech to text or voice recognition systems~\cite{Qin19_audiomitm, Yak19_audiophys, Son21_preprocessingaudio}, although we do not test this case here. 

Beyond illustrating the typical use scenarios of $A^5$, Fig.~\ref{fig:A5_complete} also serves as a reference for the notation used in the code that we share in \url{https://github.com/NVlabs/A5}.

\subsection{Additional results}

\subsubsection{FashionMNIST}

For completeness, we test $A^5/R$ and $A^5/RC$ on FashionMINST for $\epsilon_A=0.1$, training with $\epsilon_A^C=0.1$, $\epsilon_A^R=0.1$ and $\epsilon_D=0.3$.
Table~\ref{tab:fashionmnist} reports the clean and certified errors, showing that $A^5/R$ significantly boosts the clean and especially the certified errors over the CROWN-IBP baseline; $A^5/RC$ reaches even better results.

Notice that $A^5/RC$ and 
$\ell_\infty$-dist Net+MLP~\cite{Zha21_LInf} achieve similar clean errors, while $A^5$ has better certified errors.
This suggests that the integration of the design philosophy of $\ell_\infty$-dist Net (based on $\ell_\infty$-dist neurons that implement 1-Lipschitz functions) and $A^5$ may lead to even bigger improvements in terms of certified robustness in the future.

\begin{table}[h!]
    \centering
    \resizebox{\columnwidth}{!}{
    \begin{tabular}{cc|c}
           Algo & Notes & Error [certified error]\\
           \hline
           CROWN-IBP\cite{crownibp_Zha20} & From~\cite{Zha21_LInf} & 15.69\% [21.99\%]\\
           CROWN-IBP\cite{crownibp_Zha20} & Ours, $\epsilon_A^C=0.1$ & 15.11\% [22.92\%] \\
           $\ell_\infty$-dist Net+MLP~\cite{Zha21_LInf} & & 12.09\% [20.77\%]\\
           \hline
           $A^5/R$ & $\epsilon_A^R=0.1, \epsilon_D=0.3$ & 14.27\% [15.75\%]\\
           $A^5/RC$ & $\epsilon_A^R=0.1, \epsilon_D=0.3$ & 11.41\% [15.53\%]
    \end{tabular}}
    \caption{Error and certified errors on FashionMNIST, $\epsilon_A=0.1$.}
    \label{tab:fashionmnist}
\end{table}

\subsubsection{Tinyimagent}

\begin{figure}[]
    \begin{center}
    \includegraphics[width=0.47\textwidth, clip=True, trim={5.3cm 14cm 14.5cm 4cm}]{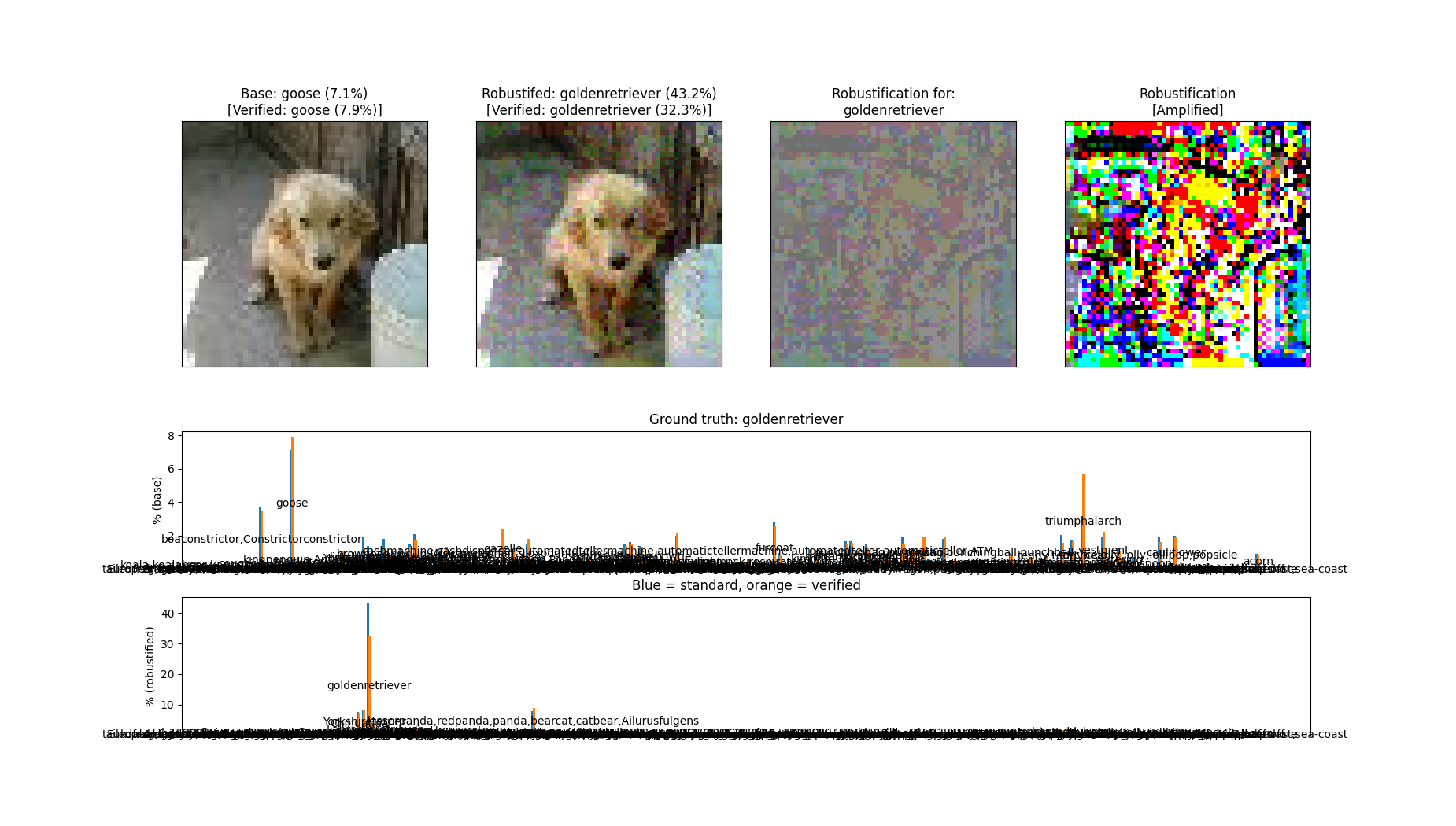} \\
    \includegraphics[width=0.47\textwidth, clip=True, trim={5.3cm 14cm 14.5cm 4cm}]{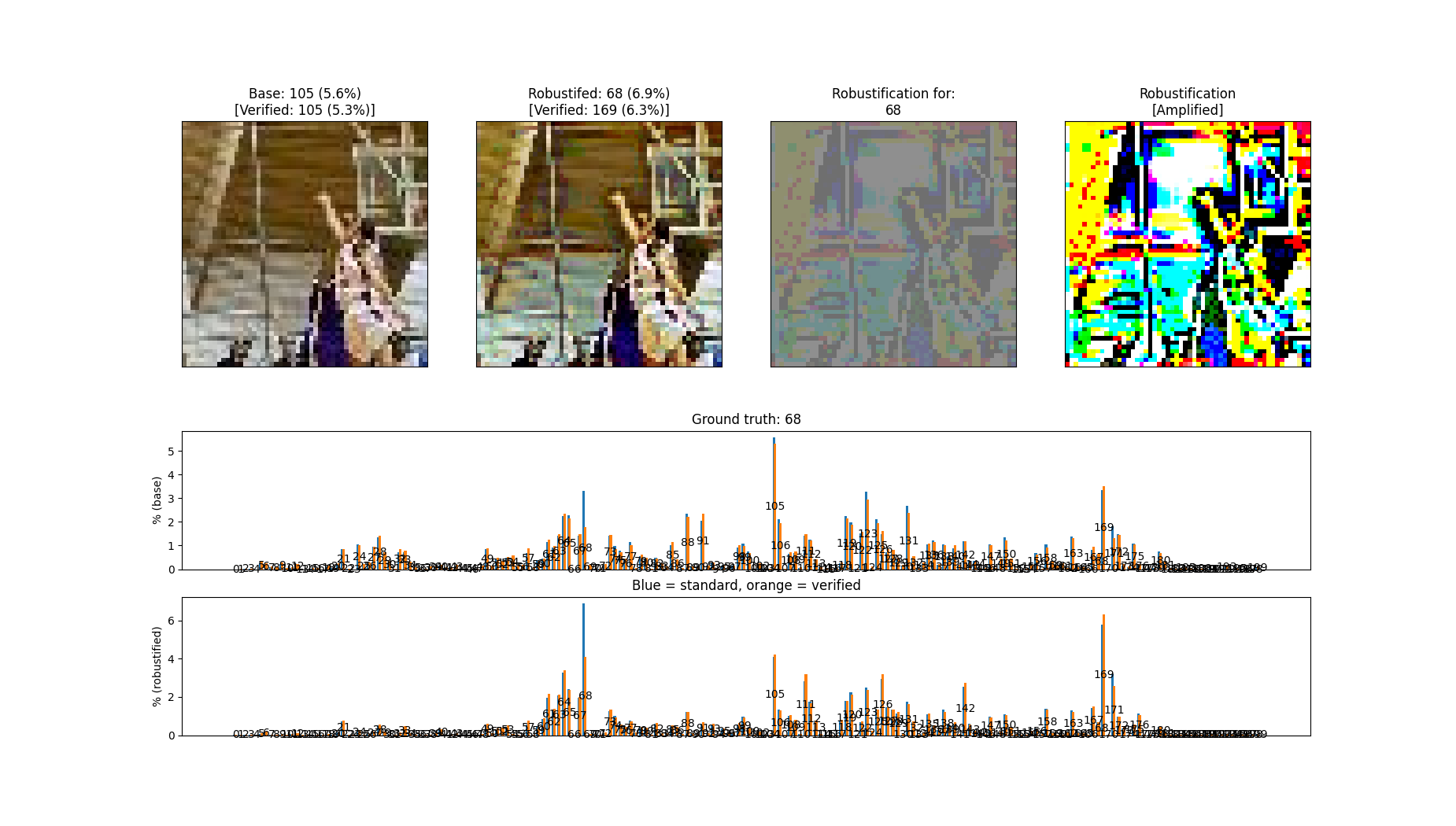}    
    \end{center}
    \caption{Robustification on tinyimagenet with $A^5/O$ (first row) and $A^5/RC$, for $\epsilon_D = 16/255$. The left panel is the original image, the central one is robustified with the augmentation signal in the right panel.}
    \label{fig:tinyimagenet}
\end{figure}

Large dataset are challenging for \emph{all} the existing defenses. On TinyImageNet we got large improvements for $A^5/O$, smaller ones for $A^5/RC$ (see Table~\ref{tab:tinyimagenet}) after 3 days of training, while the loss is still decreasing.
We beat the state-of-the-art, but we believe that the classification landscape of $C$ is not regular enough: $A^5$ exploits the gradient to find \emph{ad hoc} solutions with class specific patterns (see Fig.~\ref{fig:tinyimagenet}, compare with the more regular defensive perturbations in Fig.~\ref{fig:examples}). This is indicative of the fact that finding a general, regular defensive signal is hard for the robustifier $R$.
More research is needed to handle large datasets and leverage $A^5$ at best.

\begin{table}
\centering
    \begin{tabular}{cc}
        Method & Error, $\epsilon_D = 16 / 255$ \\
        \midrule
        CROWN-IBP & 75.33 [80.39 / 84.80] \\
        \midrule
        $A^5/O$ &  18.14 [21.24 / 25.08] \\ 
        $A^5/RC$ & 71.50 [75.19 / 78.54]
    \end{tabular}
    \vspace{-2mm}
    \caption{Results on Tinyimagenet, for $\epsilon_A^C = 1/255$. We report the clean error [PGD error / verified error].}
    \label{tab:tinyimagenet}
\end{table}

\subsection{Training recipes and network architectures}

\subsubsection{MNIST, classifier architecture}

For the MNIST classifier $C$, we adopt a neural network with the \emph{very large} architecture described in the IBP~\cite{ibp_Gow19} and the CROWN-IBP~\cite{crownibp_Zha20} papers that is also implemented (and shared) in auto{\_}LiRPA~\cite{lirpa_Xu20,autolirpa_Xu21}.

\subsubsection{MNIST, classifier training with CROWN-IBP}

For training our CROWN-IBP classifiers, we follow the instructions reported in the CROWN-IBP~\cite{crownibp_Zha20} paper, apart from some minor details like the scheduling of the learning rate or the annealing of the training attack magnitude that we change as follows.

For training attack magnitudes $\epsilon_A^C < 0.3$, we train with batch size 100 for 100 epochs with an initial learning rate of $0.001$, decreased at epochs 25 and 42 by a $10\times$ factor.
We use a \emph{SmoothedScheduler} (as implemented in auto{\_}LiRPA~\cite{lirpa_Xu20,autolirpa_Xu21}) with starting epoch 3, length 18 epochs and $mid=0.3$ to increase the value of $\epsilon_A^C$ during training.
The worst case margins are computed in all cases using the mixed CROWN-IBP bounds~\cite{crownibp_Zha20}.
For $\epsilon_A^C >= 0.3$, we use batch size 256 and 200 epochs, initial learning rate set to $0.0005$ and decreased by a $10\times$ factor at epochs 130 and 190, and a \emph{SmoothedScheduler} to increase $\epsilon_A^C$ starting at epoch 10, lasting for 50 epochs and with $mid=0.3$. We use RMSprop as optimizer.

\subsubsection{MNIST, robustifier architecture and $A^5$ training schedules}

Our robustifier $R$ for MNIST is a tiny convolutional network that takes a $1 \times 28 \times 28$ image in input, processes it with a convolutional layer with $32~~3 \times 3$ filters with stride $1$ and padding $1$, a ReLU, and a second convolutional layer with $1~~5 \times 5$ with stride $1$ and padding $2$ filter to produce the single channel, $28 \times 28$ output $\boldsymbol{z}$.

To train $R$ with $A^5/R$ on MNIST, given a pretrained (robust) classifier $C$, we use 50 epochs, batch size $256$, initial learning rate $0.0005$, whereas $\epsilon_A^R$ and $\epsilon_D$ are kept constant during the entire training process.

To fine tune $R$ and $C$ with $A^5/RC$ on MNIST, given a pretrained (robust) classifier $C$ and a pretrained roustifier $R$, we use 100 epochs
 and a fixed learning rate equal to $0.0005$; $\epsilon_A^R$ and $\epsilon_D$ are kept constant during the entire training process.
The batch size is $512$ when data augmentation is used, $256$ otherwise (see~\ref{sec:mnist_augmentation}).

In all cases, we use RMSprop as optimizer.

\subsubsection{Augmentation helps robustification by preventing overfitting}
\label{sec:mnist_augmentation}

In the case of MNIST, $A^5/RC$, data augmentation (image random shifts by max 4 pixels and random rotation by max 10 degreees) leads to better robustification performances.
Table~\ref{tab:mnist_a5rc} shows the clean, autoattack and certified errors for $A^5/RC$ with no augmentation on MNIST; these metrics are clearly inferior to the ones reported in Table~\ref{tab:mnist_a5rc_augment}, showing the importance of using data augmentation.

\begin{table}
\center
\resizebox{\columnwidth}{!}{\begin{tabular}{cc|ccccc}
$\epsilon_A^R = \epsilon_A$ & $\epsilon_D$ & $\epsilon_A^C = 0.0$ & $\epsilon_A^C = 0.1$ & $\epsilon_A^C = 0.2$ & $\epsilon_A^C = 0.3$ & $\epsilon_A^C = 0.4$ \\
\midrule
0.1 & 0.05 & 88.65 [88.65, 88.67] & 1.05 [1.88, 4.00] & 1.08 [1.62, 2.28] & 1.21 [1.93, 2.76] & 1.58 [2.28, 2.88] \\
0.1 & 0.10 & 88.65 [88.65, 88.68] & 1.07 [1.59, 3.20] & 1.06 [1.44, 1.82] & 1.03 [1.70, 2.24] & 1.53 [2.13, 2.45] \\
0.1 & 0.20 & 88.65 [88.65, 88.66] & 1.09 [1.43, 2.32] & 1.01 [1.34, 1.64] & 1.00 [1.44, 1.83] & 1.27 [1.62, 2.06] \\
0.1 & 0.30 & 88.65 [88.65, 88.65] & 1.01 [1.31, 2.10] & 0.99 [1.23, 1.51] & 1.03 [1.29, 1.59] & 1.25 [1.53, 1.88] \\
0.1 & 0.40 & 88.65 [88.65, 88.65] & 1.00 [1.30, 1.94] & 0.98 [1.13, 1.44] & 1.01 [1.24, 1.54] & 1.30 [1.47, 1.93] \\
\midrule
0.3 & 0.05 & 88.65 [88.65, 88.67] & 3.43 [9.24, 12.41] & 1.48 [5.77, 8.29] & 1.63 [5.63, 7.78] & 1.93 [5.53, 6.78] \\
0.3 & 0.10 & 88.65 [88.65, 88.66] & 1.74 [5.44, 8.15] & 1.29 [4.13, 6.46] & 1.49 [4.23, 6.19] & 2.09 [4.50, 5.32] \\
0.3 & 0.20 & 88.65 [88.65, 88.65] & 0.98 [2.26, 5.78] & 1.23 [2.34, 4.03] & 1.39 [2.55, 3.96] & 1.71 [3.03, 3.77] \\
0.3 & 0.30 & 88.65 [88.65, 88.66] & 0.94 [1.49, 3.45] & 1.11 [1.54, 2.44] & 1.40 [2.18, 3.02] & 1.56 [2.39, 2.98] \\
0.3 & 0.40 & 88.65 [88.65, 88.65] & 0.94 [1.26, 2.41] & 1.03 [1.27, 1.90] & 1.15 [1.84, 2.70] & 1.54 [2.02, 2.58] \\
\end{tabular}}
\caption{Error on clean data and (within brackets) under autoattack~\cite{autoattack_1, autoattack_2} and verified, for MNIST, $A^5/RC$, under attack $\epsilon_A = \{0.1, 0.3\}$. During training we use $\epsilon_A^R = \epsilon_A$, and no data augmentation. Comparison with Table 5 in the main paper demonstrates the benefit of using data augmentation.}
\label{tab:mnist_a5rc}
\end{table}

Our interpretation of this result is that augmentation prevents overfitting, that is significant at least in case of MNIST where errors are small.
This is further confirmed by the analysis of the training curves, where we observe the training error decreasing while validation goes up during training with $A^5/RC$ without any data augmentation.
The benefit of using data augmentation to train an MNIST robust classifier through CROWN-IBP or when using $A^5/R$ is inferior or not present at all, probably because these algorithms achieve larger errors and overfitting is less likely.

\subsubsection{CIFAR10, classifier architecture}

For the CIFAR10 classifier $C$, we adopt a neural network with the \emph{very large model} architecture described in the IBP~\cite{ibp_Gow19} and the CROWN-IBP~\cite{crownibp_Zha20} papers and implemented in auto{\_}LiRPA~\cite{lirpa_Xu20,autolirpa_Xu21}.
We do not train our own classifier with CROWN-IBP --- instead, we use the pretrained model distributed in~\url{https://github.com/huanzhang12/CROWN-IBP}. 

\subsubsection{CIFAR10, robustifier architecture and $A^5$ training schedules}

Our robustifier $R$ for CIFAR10 is a small convolutional DNN with 3 convolutional layers with $64 \times 5 \times 5$ filters with stride 1, padding 2, $64 \times 5 \times 5$ filters with stride 1, padding 2, and $3 \times 5 \times 5$ filters with stride 1, padding 2, with ReLU activations between them.
The robustifier takes a $3 \times 32 \times 32$ image in input and produces in output a vector $\boldsymbol{z}$ of the same size.

To train $R$ on CIFAR10, given the aforementioned pretrained (robust) classifier $C$, we use 100 epochs, batch size 256, learning rate $0.0005$, whereas $\epsilon_A^R$ and $\epsilon_D$ are kept constant during the entire training process.

To fine tune $R$ and $C$ with $A^5/RC$ on CIFAR10, given a pretrained (robust) classifier $C$ and a pretrained robustifier $R$, we use 200 epochs, batch size 512, learning rate $0.0005$, whereas $\epsilon_A^R$ and $\epsilon_D$ are kept constant during the entire training process.

In both cases, we use the standard data augmentation procedure for CIFAR10 images (random horizontal flips, random crop by 4 pixels max, image normalization), and RMSProp as optimizer.

\subsubsection{Additional notes for training with $A^5/RC$}

Setting the right learning rate is fundamental for $A^5/RC$.
If the learning rate is too small, the $C$/$R$ pair do not move much away from the initial condition, falling into a local minimum of the cost function and achieving little gain with respect to the optimal $R$ computed for the base $C$. If the learning rate is too large, training is unstable.
With the right learning rate, we observe a small, initial decrease of the performances of the $C$/$R$ pair, which we believe to be associated with exiting from the local minimum, but training later progresses towards much better clean and verified error rates.
To train with $A^5/RC$, we also found important to use a sufficiently large batch size to guarantee the stability of the process.

Some of the results reported in the main paper suffer indeed from a sub-optimal choice of the learning rate (and potentially other training parameters). For instance, $A^5/RC$ on CIFAR10 achieves far from optimal results for small values of $\epsilon_A^C$.
For simplicity (adoption of the same training parameters for all the metrics in the same Table) we decided to leave these metrics in the Tables without fine tuning, since the configurations affected by this issue are not the optimal ones.
Results can clearly improve after better training parameter tuning, but the presence of these metrics in the Table highlights the importance of a good choice of the training parameters.

\subsubsection{FashionMNIST}

For FashionMNIST, we use the same architectures and recipies used for $C$ and $R$ in MNIST.

\subsubsection{Tinyimagenet}

For Tinyimagenet, we start from the legacy classifier trained with CROWN-IBP that can be downloaded following the instructions provided in \url{https://github.com/Verified-Intelligence/auto_LiRPA/blob/master/doc/src/examples.md#certified-adversarial-defense-on-downscaled-imagenet-and-tinyimagenet-with-loss-fusion}. 
To train the robustifier $R$ and fine tune the classifier $C$, we use 1500 epochs, batch size 16, initial learning rate 0.0001 multiplied by a $0.8\times$ factor at epochs 500, 750, 1000, and 1250, $\epsilon_A^R = 1.1/255$ and $\epsilon_D = 16/255$.

\subsubsection{Optical Character Recognition, classifier architecture}

For OCR, we use a classifier $C$ with 4 convolutional layers ($64 \times 5 \times 5$ filters with stride 2, padding 2, $32 \times 5 \times 5$ filters with stride 2, padding 2, $16 \times 3 \times 3$ filters with stride 2, padding 1, $8 \times 3 \times 3$ filters  with stride 2, padding 1) with ReLU activations, followed by 1 fully connected layer with 512 features in input and output, ReLU activations and a last fully connected layer with 512 features in input and 62 in output.
The input of $C$ is a $3 \times 128 \times 128$ image, whereas the output is the vector of 62 logit values.

\subsubsection{Optical Character Recognition, classifier training with CROWN-IBP}

To train a traditional, non robust classifier $C$, we perform 1666 update steps using RMSprop as optimizer.
In each update step we process a batch with 672 characters (see Fig.~\ref{fig:ocr}).
The learning rate is initially set to 0.001 and decreased by a factor $10\times$ at steps 833 and 1250.
Training images $\boldsymbol{x}$ are obtained by simulating the acquisition of the (robust) characters (first row in Fig.~\ref{fig:ocr}) as $\boldsymbol{x} + \delta \boldsymbol{x}_D = A(\boldsymbol{w}+\delta\boldsymbol{w}_D)$\footnote{Notice that $\delta \boldsymbol{x}_D$ and $\delta \boldsymbol{w}_D$ are null when training a standard or a robust CROWN-IBP classifier.}, where $A$ includes: a random crop of max 5 pixels; a random rotation of max 5 degrees; a random perspective distortion with max distortion scale 0.25 (accordingly to the pytorch implementation of \emph{transforms.RandomPerspective}); adding white Gaussian noise with standard deviation in a random range from 0.2/255 to 25/255; a random blur with $\sigma$ in the range $[0.01, 1.0]$ pixels; and color jittering (with brightness, contrast, saturation and hue parameters equal to 0.1, accordingly to the pytorch implementation of \emph{transforms.ColorJitter}).

To train a robust classifier $C$, we initialize the weights of $C$ with those of the previously trained non robust classifier, then we use CROWN-IBP for an additional 2500 training steps.
Each training step uses a batch of size 128.
We use a learning rate $0.0005$ that is decreased by a factor $10\times$ after 1500 and 2000 steps, and increase $\epsilon_A^C$ from 0.0 to 0.2 in 2000 training steps, using a \emph{SmoothedScheduler} (as implemented in auto{\_}LiRPA~\cite{lirpa_Xu20,autolirpa_Xu21}) with $mid=0.3$.

\subsubsection{Optical Character Recognition, $A^5$ training schedules}

To compute the robustified characters $\boldsymbol{w}+\delta \boldsymbol{w}_D$ wth $A^5/P$, we start from a classifier $C$ trained with CROWN-IBP. 
We run $A^5/P$ for 2500 epochs, where in each epoch we process 10 batches each of size 6. 
We use a learning rate $0.005$ that is decreased by a factor $10\times$ after 1500 and 2000 epochs, whereas 
$\epsilon_A^C$ is fixed to 0.125 for the entire training process.
We use $\epsilon_D=1$ to allow the pixels to be changed from black to white (or viceversa).

To train with $A^5/PC$, we start from a classifier $C$ trained with CROWN-IBP. 
We run $A^5/P$ for 10500 epochs, where in each epoch we process 10 batches each of size 6. 
We use a learning rate $0.005$ that is decreased by a factor $10\times$ after 6000 and 8000 epochs, whereas 
$\epsilon_A^C$ is fixed to 0.125 for the entire training process.
We use $\epsilon_D=1$ to allow the pixels to be changed from black to white (or viceversa).

As the reader may have noticed, the adoption of the correct training schedule is generally important to guarantee the convergence of $A^5$ to an effective solution.

\subsection{Effect of robustification on image quality}

In many applications, the raw output of the acquisition device (and therefore its visual appearance) is not of interest for the final user --- this is for instance the case of a classifier $C$ employed in an automotive or robotic vision system.
On the other hand, for those applications where the images in output from the acquisition device are potentially consumable by human observers, it is legit asking if and how image robustification through $A^5/O$, $A^5/R$, and $A^5/RC$ affects the image quality.
$A^5$ is already created by design such that the L$_\infty$ norm of the defensive perturbation does not exceed $\epsilon_D$ (\ie, $||\delta \boldsymbol{x}_D||_\infty < \epsilon_D$), thus the user can modulate the worst case image degradation by setting $\epsilon_D$.
However, we do not know how deeply $A^5/O$, $A^5/R$, and $A^5/RC$ leverage the available perturbation space while computing $\delta \boldsymbol{x}_D$.
To quantify it, we measure the average Peak Signal to Noise Ratio (PSNR) between the original  $\boldsymbol{x}$ and the robustified $\boldsymbol{x}+\delta\boldsymbol{x}_D$ image for the CIFAR10 experiment reported in Table 2 in the main paper.
The PSNR values are reported in Table~\ref{tab:psnrs}.

\begin{table}[]
    \centering
    \resizebox{\columnwidth}{!}{
    \begin{tabular}{c|cc|ccc}
    Algo & $\epsilon_D$ & Worst case & $\epsilon_A^R=4/255$ & $\epsilon_A^R=8/255$ & $\epsilon_A^R=16/255$ \\
    \hline
\multirow{5}{*}{\rotatebox{90}{$A^5/O$}}
         & 4/255 &36.09dB & 47.04 & 46.40 & 44.88 \\
         & 8/255 & 30.07dB & 40.74 & 39.95 & 38.51 \\
         & 16/255 & 24.05dB & 34.56 & 33.90 & 32.26 \\
         & 32/255 & 18.03dB & 28.29 & 27.74 & 26.12 \\
\hline
\multirow{5}{*}{\rotatebox{90}{$A^5/R$}}
         & 4/255 & 36.09dB & 36.27 & 36.27 & 36.35 \\
         & 8/255 & 30.07dB & 30.31 & 30.30 & 30.35\\
         & 16/255 & 24.05dB & 24.47 & 24.40 & 24.45 \\
         & 32/255 & 18.03dB & 18.75 & 18.70 & 18.61\\
\hline
\multirow{5}{*}{\rotatebox{90}{$A^5/RC$}}
         & 4/255 & 36.09dB & 36.27 & 36.27 & 36.35\\
         & 8/255 & 30.07dB & 30.27 & 30.27 &  30.35\\
         & 16/255 & 24.05dB & 24.34 & 24.34 & 24.34\\
         & 32/255 & 18.03dB & 18.57 & 18.56 & 18.52\\
\hline
    \end{tabular}}
    \caption{Average PSNR of the robustified $\boldsymbol{x}+\delta\boldsymbol{x}_D$ with respect to the vanilla $\boldsymbol{x}$ for CIFAR10 images and robustification with $A^5/O$, $A^5/R$, and $A^5/RC$ and different settings (the same used for Table 2 in the main paper). The PSNR reported in the third column represents the worst case, where each pixel value is either increased or decreased by $\epsilon_D$ exactly.}
    \label{tab:psnrs}
\end{table}

This Table highlights a difference in the images robustified by $A^5/O$ and those output by $A^5/R$ and $A^5/RC$.
In this first case, the image degradation is proportional to $\epsilon_D$, as expected, but it is far from the worst case.
In other words, $A^5/O$ identifies local optima where many pixel values are changed by less than $\epsilon_D$.
On the other hand, both $A^5/R$ and $A^5/RC$ are characterized by PSNR values that are close to the worst case ones, meaning that the pixel values are often changed by either $+\epsilon_D$ or $-\epsilon_D$.
Such observation is consistent with the images shown in Fig.~\ref{fig:examples}, where $\delta \boldsymbol{x}_D$ shows a sparse structure for $A^5/O$, while it is characterized by large, constant value areas in case of $A^5/R$ and $A^5/RC$.

Overall, $A^5/RC$ provides a more effective robustification than  $A^5/O$; this is achieved at the cost of a larger image degradation, anyway controlled by $\epsilon_D$.

\subsection{Resources}

For training and testing $A^5$ we use either an HP Z8 G4 Workstation equipped with an Intel Xeon Gold 6128 @ 3.40GHz CPUs, RAM 48G and two NVIDIA GeForce 2080 Ti GPUs, each with 12G RAM, or an NVIDIA DGX-1 equipped with Intel Xeon E5-2698 v4 @ 2.20GHz CPUs, RAM 48G, and 8 NVIDIA  Tesla V100-SXM2 GPUs, each with 32G RAM.
All training and testing are done using a single GPU, running multiple experiments in parallel on the same machine when possible. Typical training times are in the order of few hours to for robust CROWN-IBP classifiers on MNIST, CIFAR10, FashionMNIST; $A^5/O$, $A^5/R$ and $A^5/RC$ take up to one day of training on the same data, while training on Tinyimagenet requires multiple days. Typical training times for $A^5/P$ and $A^5/PC$ on the 62 characters dataset are again in the order of few hours.

%%%%%%%%% REFERENCES
{\small
\bibliographystyle{ieee_fullname}
\bibliography{aaaaa}
}

%\newpage
%\input{sections/supplementary.tex}

%\newpage
%\input{sections/related_work_extended}

\end{document}